\documentclass[review]{elsarticle}

\usepackage{lineno,hyperref}
\usepackage{amsfonts}
\usepackage{subfigure}
\usepackage{graphicx}
\usepackage{amsmath}
\usepackage{amssymb}
\usepackage{times}
\usepackage{algorithmic}
\usepackage{algorithm}
\usepackage{booktabs}
\usepackage{url}
\usepackage{multirow}
\usepackage{color}

\journal{Medical Image Analysis}

\newcommand{\KARE}{\mathrm{KARE}}
\newcommand{\KRRE}{\mathrm{KRRE}}

\begin{document}

\begin{frontmatter}

\title{A Divide-and-Conquer Approach to\\ Compressed Sensing MRI}

\author{Liyan Sun, Zhiwen Fan, Xinghao Ding, Congbo Cai, Yue Huang, John Paisley$^{\dagger}$}
\address{Fujian   Key   Laboratory   of   Sensing   and   Computing   for   Smart   City,  Xiamen   University, Fujian, China\\
$^{\dagger}$Department of Electrical Engineering, Columbia University, New York, NY, USA\\
Declarations of interest: none}

\begin{abstract}
Compressed sensing (CS) theory assures us that we can accurately reconstruct magnetic resonance images using fewer k-space measurements than the Nyquist sampling rate requires. In traditional CS-MRI inversion methods, the fact that the energy within the Fourier measurement domain is distributed non-uniformly is often neglected during reconstruction. As a result, more densely sampled low-frequency information tends to dominate penalization schemes for reconstructing MRI at the expense of high-frequency details. In this paper, we propose a new framework for CS-MRI inversion in which we decompose the observed k-space data into ``subspaces'' via sets of filters in a lossless way, and reconstruct the images in these various spaces individually using off-the-shelf algorithms. We then fuse the results to obtain the final reconstruction. In this way we are able to focus reconstruction on frequency information within the entire k-space more equally, preserving both high and low frequency details. We demonstrate that
the proposed framework is competitive with state-of-the-art methods in CS-MRI in terms of quantitative performance, and often improves an algorithm's results qualitatively compared with its direct application to k-space.
\end{abstract}


\end{frontmatter}


\section{Introduction}

MRI is an important medical imaging technique because of its harmlessness and the high resolution information it measures in soft tissue. The data acquisition process directly measures Fourier coefficients of the object being imaged, which can then be recovered by an inverse Fourier transformation. A significant drawback of MR imaging is that data acquisition is relatively slow, meaning that a patient has to remain still for a long time to avoid producing motion artifacts. This is something especially difficult for children and those who are critically ill. Thus, accelerating the measurement speed while maintaining a diagnostic-quality reconstruction is a major challenge in MR imaging.

Compressive sensing (CS) theory \cite{1,2} has shown that it is possible to accurately reconstruct MR images with significantly fewer measurements than mandated by the Nyquist sampling theorem. The typical approach to CS-MRI can be generalized as the optimization problem
\begin{equation}\label{eq1}
x =  \arg\min_x \| F_u x - y \|_2^2 + f_{\theta}(x).
\end{equation}
The goal is to reconstruct the vector $x \in\mathbb{C}^{N}$, which corresponds to values in the MR image $X \in {\mathbb{C}^{\sqrt N  \times \sqrt N}}$.
The matrix ${F_u} \in {\mathbb{C}^{u \times N}},u < N$ is an under-sampled Fourier basis used to directly measure the k-space data $ y\in\mathbb{C}^{u}$. In this objective, the first term enforces that the Fourier coefficients of $x$ agree with the observations $y$. Since many vectors $x$ will satisfy this requirement, the penalty $f_{\theta}(x)$ taking parameters $\theta$ searches for an $x$ with additional desired properties such as smoothness.

\subsection{Related Work}
Many CS-MRI algorithms have been proposed in the framework of Equation (\ref{eq1}). These developments mostly fall into two categories depending on whether they focus on new objective functions or on new efficient optimization algorithms.

Along the first line, new objective functions for CS-MRI exploit the sparsity of MR images under different transform domains. Although medical images may not be sparse in the image domain, they can be projected onto a transform basis incoherent with the Fourier basis where they show a high degree of sparsity. Such bases include wavelets \cite{3}, total variation \cite{3,4,5}, contourlets \cite{6}, Walsh \cite{7} and PCA \cite{8,9,10}. Patch-based bases include directional wavelets (PBDWS) \cite{30,31}, a graph-based redundant wavelet transform (GBRWT) \cite{37}, and dictionary bases constructed \textit{in situ} using dictionary learning algorithms like KSVD \cite{11} and BPFA \cite{12}. As researches in deep learning methods thrive, the popular deep learning architectures are brought into CS-MRI \cite{38,39,40}. In recent years, other approaches have enforced structural sparsity \cite{13,14,41}, nonlocal priors \cite{15, 16}, and approximations to the desired $\ell_0$ penalty such as the convex $\ell_1$,
FOCUSS, $\ell_p$ (for $0<p<1$), IRLS and smoothing functions \cite{17,18,19,20}.

One particular method that we highlight and will use is the patch-based directional wavelet (PBDWS) \cite {30,31}. This CS-MRI objective function assumes that patches extracted and vectorized from the reconstructed MRI are sparse in the Haar wavelet domain. A key novelty is that each patch is vectorized in a way that depends on the geometric structure of the signal in that patch, and is chosen such that sparsity is maximized. In this way image details can be preserved better while satisfying the need for sparsity.

Another line of work has been devoted to finding more efficient ways to optimize the various objective functions arising from the framework of Equation (\ref{eq1}), such as TVCMRI \cite {21}, RecPF {\cite{22}}, FISTA \cite {23}, FCSA \cite {24}, pFISTA \cite{44} (an algorithmic variation on SparseMRI \cite{3}), Bregman \cite {25} and ADMM \cite {12}. These methods typically work by representing the objective function in a way that is easier to optimize iteratively.

\subsection{Our contribution}

\begin{figure}\centering
\subfigure[Brain MRI]{\label {figure1a}
\includegraphics[height=2in]{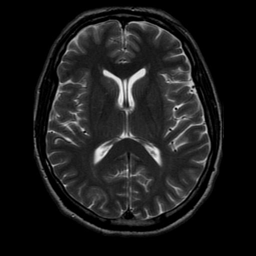}}\qquad
\subfigure[k-space coefficients]{\label {figure1b}
\includegraphics[height=2in]{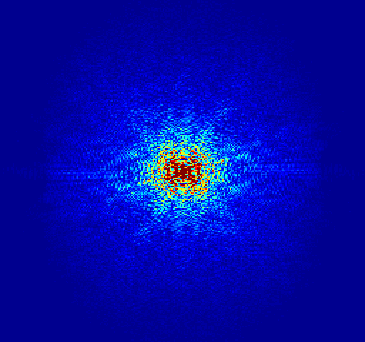}}
\caption{k-space corresponding to a brain MRI.}
\label{figure1}
\end{figure}

In general, k-space data of MRI are not distributed uniformly across all frequency bands in energy and magnitude. As an example, we show the magnitudes of k-space data for an MRI of the brain in Figure \ref{figure1}. As is clearly evident, the energy is concentrated much more in the low frequency region of k-space than the high frequency region. This is a well-known fact about MRI, which is considered in virtually all compressed sensing frameworks through the design of variable density sampling masks that sample more heavily in the low frequency part of k-space to ensure the basic structure is measured. However, as a result of using a squared error penalty term of the form $\left\| {{F_u}x - y} \right\|_2^2$ as in the Equation (\ref{eq1}), CS-MRI inversion will result in these low frequency measurements dominating the reconstruction of $x$. Since high frequency coefficients encode structural details such as edges and curves, the accumulation of error caused by overlooking high frequency information may not
give the detail needed for diagnosis. In this paper we propose a general framework
for reconstructing MRI measured with compressed sensing in a way that focuses more equally on all regions of k-space, and can be incorporated into any existing reconstruction algorithm. In particular, our approach breaks down the data fidelity squared error into various parts in a ``divide-and-conquer'' (DAC) manner and can incorporate \textit{any} existing CS-MRI inversion algorithm to reconstruct those parts.

The paper is organized as follows: In Section \uppercase\expandafter{\romannumeral2}, we discuss the non-uniformity property of the k-space data and its relationship to the reconstruction accuracy for CS-MRI. To address this issue, we propose a three-step divide-and-conquer (DAC) framework in Section \uppercase\expandafter{\romannumeral3}. Section \uppercase\expandafter{\romannumeral4} demonstrates the performance of our DAC framework on several MR data with different under-sampling masks and ratios. We provide further discussion on parameter setting and noise characteristics in Section \uppercase\expandafter{\romannumeral5}.

\subsection{Notation}

We use the following notation: the fully-sampled or reference MRI is denoted as $X_f \in {\mathbb{C}^{\sqrt N  \times \sqrt N}}$ with the vectorized form $x_f \in\mathbb{C}^{N}$. Upper case means the representation in 2D while lower case means a vectorized form of the corresponding 2D representation. The subscript $f$ means fully-sampled. Projecting these into the frequency domain, we represent the corresponding projections as $Y_f \in {\mathbb{C}^{\sqrt N  \times \sqrt N}}$ and $y_f \in\mathbb{C}^{N}$. The measured k-space $Y \in {\mathbb{C}^{\sqrt N  \times \sqrt N}}$ has its unsampled positions padded with zeros, but the vectorized form $y \in\mathbb{C}^{u}$ will only have the sampled locations. The vector $y$ can be obtained by multiplying $x_f$ by an under-sampled Fourier basis matrix $F_u$. We further denote the reconstructed MRI image as $X_{r} \in {\mathbb{C}^{\sqrt N  \times \sqrt N}}$ and its k-space representation as $Y_{r} \in {\mathbb{C}^{\sqrt N  \times \sqrt N}}$. If we define a filter $H$ with limited spatial support, which is usually much smaller than the convolved ${\sqrt N  \times \sqrt N}$ image, the block Toeplitz matrix of the filter $H$ is defined as $\widetilde H \in {\mathbb{R}^{\sqrt N  \times \sqrt N}}$. Furthermore, the frequency response of filter $h$ is defined as $\widehat H \in {\mathbb{R}^{\sqrt N  \times \sqrt N}}$.

\section{Motivation: Nonuniform k-space density}\label{sec.motivation}

K-space data is known to be distributed non-uniformly in energy, as illustrated with the brain MRI in Figure \ref{figure1}. While k-space magnitudes tend to be larger in low frequency bands, much diagnostically important detail information is known to be in the high frequency regions. However, many existing CS-MRI methods treat all errors equally, which tends to favor high-magnitude, low-frequency information in reconstructions at the expense of the details.

To further analyze the phenomenon, we define the following two simple measures: the \textit{k-space absolute reconstruction error} (KARE) and the \textit{k-space relative reconstruction error} (KRRE). These are the same size as the k-space, with the $(i,j)^{th}$ element denoted as follows,
\begin{eqnarray}
\label{eq2} & \left( {\KARE} \right)_{ij} = \big|( {Y_r} )_{ij} - (Y_f )_{ij} \big|\\
\label{eq3} & \left( {\KRRE} \right)_{ij} = \big|( {Y_r} )_{ij} - (Y_f )_{ij} \big|}/{\big| ( {Y_f} )_{ij} \big|.
\end{eqnarray}
Here, ${Y_f} \in {\mathbb{C}^{\sqrt N  \times \sqrt N }}$ is the fully-sampled k-space data, and ${ {Y_r} } \in {\mathbb{C}^{\sqrt N  \times \sqrt N }}$ is the k-space of the MRI reconstructed from subsamples.


\begin{figure}[htb!]
\begin{center}
\subfigure[\footnotesize KARE map of PBDWS\qquad]{\label {figure2a}%
\includegraphics[width=0.235\textwidth]{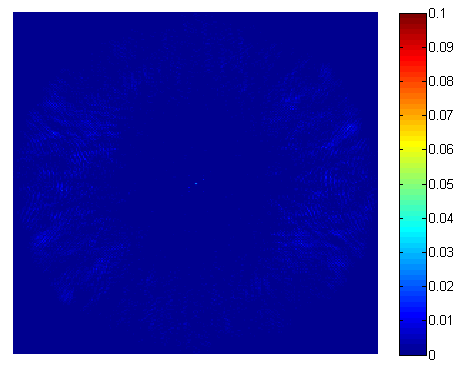}}
\subfigure[\footnotesize KRRE map of PBDWS\qquad]{\label {figure2b}%
\includegraphics[width=0.235\textwidth]{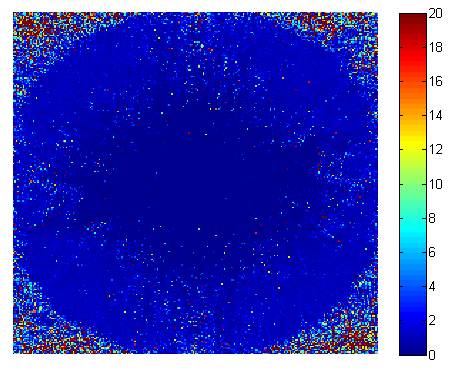}}
\subfigure[\footnotesize Redistributed KARE\qquad]{\label {figure2c}%
\includegraphics[width=0.235\textwidth]{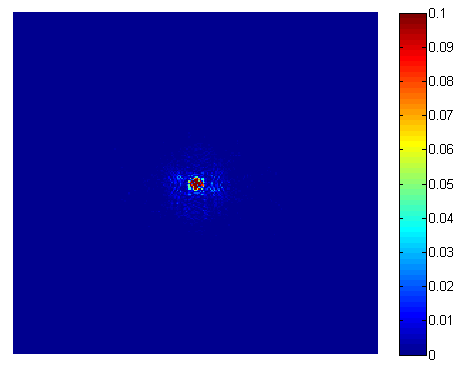}}
\subfigure[\footnotesize Redistributed KRRE\qquad]{\label {figure2d}%
\includegraphics[width=0.235\textwidth]{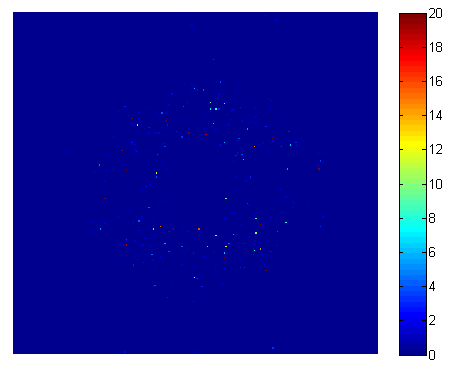}}
\subfigure[\footnotesize PBDWS (SSIM = 0.91)]{\label {figure2e}%
\includegraphics[width=0.227\textwidth]{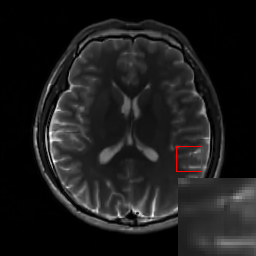}}
\subfigure[\footnotesize Redistributed (SSIM = 0.97)]{\label {figure2f}%
\includegraphics[width=0.227\textwidth]{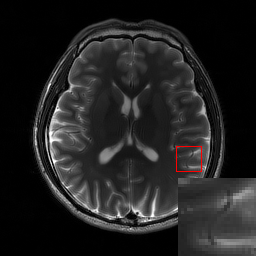}}
\caption{CS-MRI reconstruction results using PBDWS and an ad-hoc error redistribution for $10\%$ 2D random sampling. Since the ground truth is not known in real experiments, this motivating example is purely illustrative. (a) The KARE map of the PBDWS reconstruction. (b) The KRRE map of the PBDWS reconstruction. (c) The KARE map of the redistributed k-space error reconstruction. (d) The KRRE map of the redistributed k-space error reconstruction.
(e) CS-MRI reconstruction image using PBDWS. (f) CS-MRI reconstruction after error redistribution.}
\label{figure2}
\end{center}
\end{figure}

For the same brain MRI, we plot in Figure \ref{figure2} the reconstructed image and the corresponding KARE and KRRE images using a CS-MRI inversion method called PBDWS \cite{31} with $10\%$ under-sampling and a 2D random mask. In Figures \ref{figure2a} and \ref{figure2b}, we observe that data in the peripheral high frequency regions suffer almost the same absolute but larger relative errors. As an illustrative experiment under the assumption that we have access to the fully-sampled k-space, one can manually redistribute the k-space error from PBDWS more evenly in such a way that the PSNR remains unchanged, as shown in Figures \ref{figure2c} and \ref{figure2d}. However, as shown in Figures \ref{figure2e} and \ref{figure2f}, after redistribution the reconstruction has much better visual quality, and a larger structural similarity measure (SSIM).

This simulation supports the claim that better high frequency k-space reconstruction accuracy, even at the expense of low frequency information, can lead to better visual quality. However, conventional CS-MRI methods tend to leave the high frequency information deemphasized in their reconstruction objectives, coupled with heavier sub-sampling in low-frequency regions.
While in previous work such information is implicitly modeled, however, to our knowledge this problem has not been explicitly addressed for CS-MRI inversion problems.

For example, in \cite {27} the low frequency k-space data is reconstructed first because of its dense distribution, then the reconstructed low frequency portions are padded back to the measurements for the second-stage reconstruction. The performance improvement of this method is limited due to fact that the reconstruction error of the low frequency bands propagate to the later reconstruction. The high and low frequency regions are separated using the rectangle box which can cause Gibbs effects and introduce artifacts. In \cite{4,5} the horizontal and vertical differential images are reconstructed and then used as gradient constraints, where they mainly focus on the sparse nature of MRI in the gradient domain. In \cite{26}, a convolutional constraints is proposed, but the work non-uniformity in k-space. In \cite{29}, a method called HiSub CS-MRI formalized a link between the k-space and wavelet domain to apply separate under-sampling and reconstruction for high/low frequency k-space data. In the HiSub method,
 the high/low frequency regions in k-space are defined based on the separation of wavelet sub-bands; compressed sensing techniques are used for the high frequency region while parallel imaging is used for the low-frequency region. The HiSub method relies on the specific sampling pattern and is not exclusively based on CS-MRI ideas.
In \cite{28}, the local scale mixture model is proposed to decompose the MR images into dual block sparse components: total variation for piecewise smooth parts and wavelet for residuals, but the decomposition only depends on the different priors between the total variation and wavelet in spatial domain.

The non-uniformity in k-space motivates us to reconsider the CS-MRI problem using a divide-and-conquer (DAC) framework that can be implemented using existing inversion algorithms. Our method consists of three steps: subspace decomposition, subspace reconstruction and subspace integration. While the word ``subspace'' is well-defined for linear algebra, here we mean a specific frequency view into which the k-space measurements are decomposed using standard filtering techniques. This method allows for the algorithms to deal with the high and low frequency k-space data separately to better preserve fine structural details. Although the idea is simple, we note that the proposed subspace method exhibits great potential for recovering fine details in MRI by better preserving the high frequency information possibly improving the diagnosis quality of medical imaging applications.\footnote{We wish to emphasize here that our proposed method does \textit{not} simply partition the k-space data into disjoint frequency
regions.}

\section{Divide-and-conquer Subspace Framework}

\begin{figure}
\begin{center}
   {\includegraphics[width=1\columnwidth]{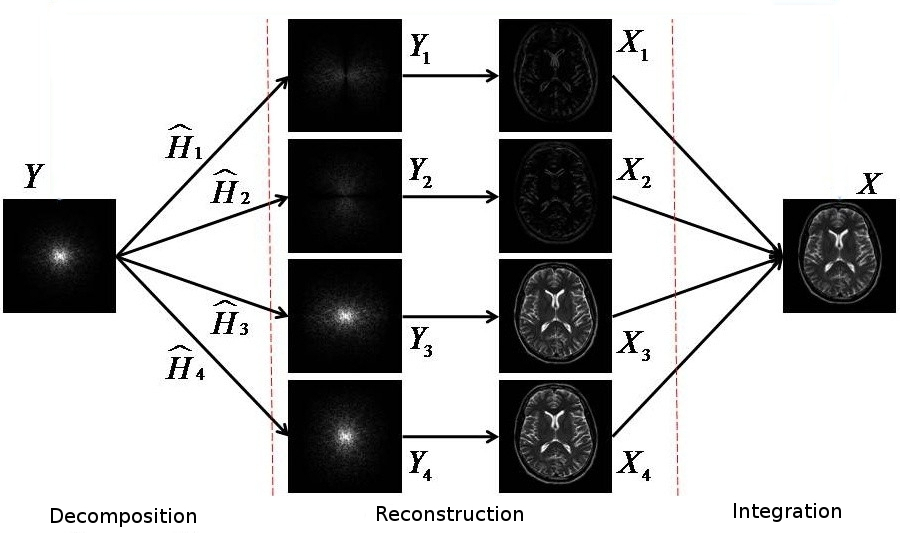}}
   \caption{Flowchart of proposed subspace method.}
\label {figure3}
\end{center}
\end{figure}

The proposed subspace method includes three steps: subspace decomposition, subspace reconstruction and subspace integration. We display a flowchart of the method in Figure \ref{figure3}. In this section we first discuss each of the three steps above separately. Then we connect these three steps to the objective function given in Equation (\ref{eq1}). At the end of the section we summarize the proposed DAC CS-MRI framework in Algorithm \ref{alg.1}. Here we adopt the HoriVert subspace decomposition for illustration, which we will elaborate in later section.

\subsection{Subspace decomposition}

In order to reconstruct the corresponding image in each subspace, we need to first define the subspaces and then obtain the partially observed k-space data in each subspace. We call this process subspace decomposition. According to signal processing theory, a lossless decomposition can be accomplished using filters. In our case, we use a set of linear filters for their simplicity. We let $\{ {{H_i}} \}_{i = 1}^S$ be the impulse responses of the set of chosen filters, where $S$ is selected in advance. In principle, a decomposition in the image domain can be formulated as follows,
\begin{eqnarray}\label{eq3}
{X_{f_i}} = X_f * {H_i},
\end{eqnarray}
where $X_{f_i} \in {\mathbb{C}^{\sqrt N  \times \sqrt N}}$ is the convolved image of the fully-sampled in the $i$th subspace and $*$ denotes the 2D convolution.

We equivalently work within the Fourier domain and obtain a series of frequency responses denoted as $\{ {{{\widehat H}_i}\in {\mathbb{C}^{\sqrt N  \times \sqrt N}} } \}_{i = 1}^S$. As is well known, matrix convolution in the image domain can now be converted into an element-wise matrix multiplication operation in the Fourier domain. Thus Equation (\ref{eq3}) is equivalent to
\begin{eqnarray}\label{eq4}
{Y_{f_i}} = Y_f \odot {\widehat H_i},
\end{eqnarray}
where ${{Y_{f_i}} \in {\mathbb{C}^{\sqrt N  \times \sqrt N}}}$ is the Fourier transformation of $X_{f_i}$ and $\odot$ denotes element-wise multiplication.

Using the under-sampling, a zero-filled partial k-space view, $ Y \in {\mathbb{C}^{\sqrt N  \times \sqrt N}}$, of ground truth $X$ can be written as
\begin{eqnarray}\label{eq5}
 Y = M \odot Y_f,
\end{eqnarray}
where $M \in {\mathbb{R}^{\sqrt N  \times \sqrt N}}$ is the under-sampling mask in which the sampled locations contain ones and unmeasured frequencies contain zeros. By simple algebra, the zero-filled, under-sampled k-space data in each subspace can now be derived as
\begin{eqnarray}\label{eq7}
\begin{array}{l}
{ Y_i} = M \odot {Y_{f_i}} = M \odot Y_f \odot {\widehat H_i} =  Y \odot {\widehat H_i}.
\end{array}
\end{eqnarray}
Equation (\ref {eq7}) indicates that the partial k-space data decomposed in each subspace can be derived via the element-wise multiplication between the frequency response of the filter and the original partially observed k-space data, or equivalently between the partially observed frequency response and the complete k-space of the MRI. We will use this latter observation in our reconstruction algorithm.

For the remaining derivation, we work with the problem by converting from matrix element-wise multiplication to matrix-vector multiplication. Thus the impulse responses $\{ {{H_i}}\}_{i = 1}^S$ of the filters is rewritten as a circulant matrix $\{ {{{\widetilde H}_i}\in {\mathbb{R}^{ N  \times  N}}}\}_{i = 1}^S$. (Note that $\widehat{H}_i$ and $\widetilde{H}_i$ differ in that the first operates on two-dimensional k-space data while the second is on vectorized images.)
An equivalent form of Equation (\ref {eq3}) and Equation (\ref {eq4}) can now be written as
\begin{eqnarray}\label{eq8}
{x_{f_i}} = \widetilde H_i x_f,
\end{eqnarray}
where $x_{f_i} \in\mathbb{C}^{N}$ is the vectorized form of $X_{f_i}$. Similarly, Equation (\ref {eq7}) can be written as
\begin{eqnarray}\label{eq9}
{y_i} = {F_u}{x_{f_i}},
\end{eqnarray}
while the original measured k-space of Equation (\ref{eq5}) is
\begin{eqnarray}\label{eq6}
 y = {F_u}x_f.
\end{eqnarray}
We next return to the filter banks considered in Section \ref{sec.filterbanks}.

\subsection{Subspace Reconstruction}\label{sec.reconstruction}

After dividing k-space into partial frequency views $y_i$, it is intuitive to exploit this isolated information and reconstruct the corresponding images in each subspace separately. This will ensure that high frequency information, captured in certain $y_i$ by appropriate filter banks, is not sacrificed in favor of the far greater number of high magnitude, low frequency measurements. To this end, we define a subspace-specific optimization problem with an appropriate regularization term to be determined. The reconstruction of each subspace can be formulated by minimizing the following objective function
\begin{eqnarray}\label{eq12}
  \mathop {\arg \min }\limits_{{x_i}} \frac{\mu_i }{2}\left\| {{F_u}{x_i} - {y_i}} \right\|_2^2 + {\rho _i}({x_i}),
\end{eqnarray}

where $\rho_i( {{x_i}})$ enforces desired properties of the reconstructed subspace image and $\mu_i$ is the regularization parameter for the data fidelity term. As with the filters, $H_i$, these penalty functions $\rho_i$ can be chosen to be any CS-MRI inversion algorithm. We note the new data fidelity squared error is proposed for better preservation for high frequency information, which also distinguish the proposed subspace method with merely regularizing the filtered subspace images with a unified data fidelity term. In this paper we use three recent state-of-the-art CS-MRI reconstruction methods: FCSA \cite{24}, WatMRI \cite{14} and PBDWS \cite{30}. To summarize, these penalties are the following:
\begin{equation}
 \rho_i(x_i) = \left\lbrace\begin{array}{ll}
                             \alpha \text{TV}(x_i) + \beta\|\Phi x_i\|_1 & \text{FCSA}\\
                              \alpha \text{TV}(x_i) + \beta\|\Phi x_i\|_{\text{tree}} & \text{WatMRI}\\
                              \|B_w x_i\|_0 & \text{PBDWS}
                           \end{array}\right.
\end{equation}

All methods use the squared error as a data fidelity term. In FCSA, the objective function is split into TV and wavelet $L_1$ regularization sub-problem in an iterative manner, where each sub-problem is solved efficiently using proximal gradient descend algorithm. Then the reconstructed MRI is obtained via the weighted average of the solutions of the two sub-problems. In this sense it is an algorithmic development of the classic SparseMRI framework \cite{3}. WatMRI instead imposes wavelet tree group sparsity on the MRI via $\Phi$. The optimization is similar to FCSA, using splitting techniques. Both FCSA and WatMRI are CS-MRI methods based on a global sparse regularization with non-adaptive transform bases. Thus they are suitable to stand for the CS-MRI methods with fixed transform basis. PBDWS on the other hand is a patch-based method in which the MRI to be reconstructed is divided into patches and wavelets are used in a way that considers the geometric structure of the patch under consideration with the
goal of maximizing sparsity. Thus PBDWS is representative of CS-MRI algorithms that use an adaptive transform basis, but the proposed framework is not limited to using methods. We choose the three methods for illustrative purposes and because they are high quality algorithms.

We experiment with these three methods to show that our method can provide general improvement to many existing CS-MRI models. In our experiments, we apply the same reconstruction algorithm to all subspaces, but with different parameter settings. If the MRI is divided into $n$ subspaces via $\widetilde{H}_1,\dots,\widetilde{H}_n$, then the chosen algorithm would be run independently $n$ times, once for each subspace. Therefore, our framework increases computation time by a factor of $n$, but the independence of each optimization allows for a straightforward parallelization.

\subsection{Subspace Integration}\label{sec.integration}

Since the subspace decomposition is a linear decomposition using linear filters, if the decomposition is complementary but not redundant then integrating these results into a final reconstruction can be done by simply adding the images together,
\begin{equation}\label{eq18}
 x = \sum_{i = 1}^N {{x_i}}.
\end{equation}
We also consider a Tikhonov regularization method by formulating the integration according to the following objective function
\begin{equation}\label{eq19}
 x = \arg \mathop {\min }\limits_x \sum\limits_{i = 1}^n {{\lambda _i}\| {{x_i} - {\widetilde {H}_i}x} \|_2^2}.
\end{equation}

This objective function admits a closed-form solution, but direct computation is infeasible because of the high dimensionality of $\widetilde H_i$.  However, by transferring the problem to the Fourier domain the reconstructed k-space is calculated element-wise followed by an inverse Fourier transform,
\begin{eqnarray}\label{eq20}
\textstyle X = F^H \big(\sum_i \lambda_i \widehat H_i^H \odot {\widehat X}_i\big)\big/\big(\sum_i \lambda_i\| {\widehat H}_i \|_2^2\big).
\end{eqnarray}
Here, the division is element-wise, as is the magnitude of ${\widehat H}_i$ in the denominator. We discuss a method for determining each $\lambda_i$ below in Section \ref{sec.overall-objective}.

\subsection{Filter banks for dividing and conquering}\label{sec.filterbanks}

We consider two kinds of filter banks in this paper: one based on the horizontal and vertical redundant filter bank (HoriVert), and another based on the Gaussian complementary filter bank (Gaussian).

\subsubsection{HoriVert subspace decomposition}

\begin{figure}
\begin{center}
   \subfigure[\scriptsize $\widehat H_{1}$]{\includegraphics[width=0.156\columnwidth]{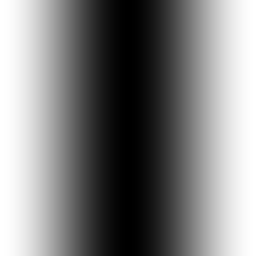}}
   \subfigure[\scriptsize $\widehat H_{3}$]{\includegraphics[width=0.156\columnwidth]{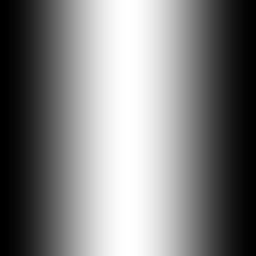}}
   \subfigure[\scriptsize $\widehat H_{2}$]{\includegraphics[width=0.156\columnwidth]{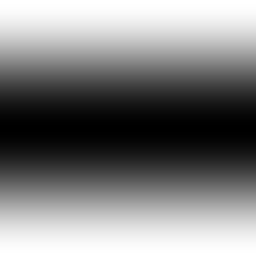}}
   \subfigure[\scriptsize $\widehat H_{4}$]{\includegraphics[width=0.156\columnwidth]{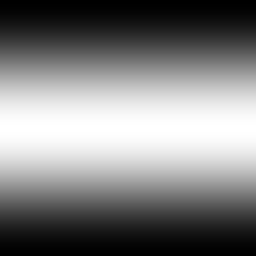}}
   \subfigure[\scriptsize $\widehat G_{lp}$]{\includegraphics[width=0.156\columnwidth]{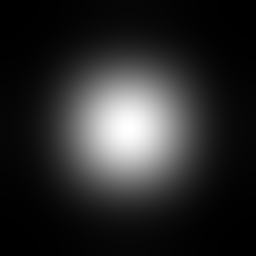}}
   \subfigure[\scriptsize $\widehat G_{hp}$]{\includegraphics[width=0.156\columnwidth]{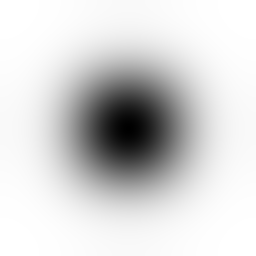}}
   \caption{The frequency response of the HoriVert redundant filter banks and the Gaussian complementary filter banks. White corresponds to one and black corresponds to zero, with a smooth transition in between.  (a)--(f) are respectively the vertical high frequency, vertical low frequency, horizontal high frequency, horizontal low frequency, Gaussian low frequency, Gaussian high frequency filters.}
\label {figure4}
\end{center}
\end{figure}

Because much of the high frequency details in MRI can be represented as vertical or horizontal edges, we consider a decomposition of k-space into horizontal/vertical high and low frequency subspaces. We adopt the following four filters for decomposition: $h_1 = [-0.5,0.5]$ and $h_3 = [0.5,0.5]$ for vertical high and low frequency subspaces, and $h_2 = [-0.5,0.5]^{T}$ and $h_4 = [0.5,0.5]^{T}$ for horizontal high and low frequency subspace. The frequency responses of these filters satisfy the relationships
\begin{eqnarray}
\label{eq9} {\widehat H_1} + {\widehat H_3} = \textbf{1}\\
\label{eq10} {\widehat H_2} + {\widehat H_4} = \textbf{1}
\end{eqnarray}
where $\textbf{1}$ is the all-ones matrix. It's easy to verify that
\begin{eqnarray}
\label{eq11} {Y_1} + {Y_3} = Y \odot {H_1} + Y \odot {H_3} = Y\left( {{H_1} + {H_3}} \right) = Y\\
\label{eq12} {Y_2} + {Y_4} = Y \odot {H_2} + Y \odot {H_4} = Y\left( {{H_2} + {H_4}} \right) = Y
\end{eqnarray}
Therefore, the proposed filtering scheme is redundant and meets the requirements for completeness, and is thus lossless. We call this proposed decomposition scheme the HoriVert subspace decomposition. We display the frequency responses of these filter banks in Figure \ref{figure4}(a)--(d).

\subsubsection{Gaussian subspace decomposition}

We also test our DAC method using spatial Gaussian filters. We design the Gaussian low-pass filter, denoted as ${g_{lp}}$, with $5\times5$ spatial support and unit standard deviation and similar for the corresponding high-pass filter, denoted ${g_{hp}}$. This gives frequency responses $\widehat{G}_{lp}$ for $g_{lp}$ and $\widehat{G}_{hp}$ for $g_{hp}$. $\widehat{G}_{lp}$ and $\widehat{G}_{hp}$ have a similar lossless property,
\begin{eqnarray}
\label{eq13} {\widehat G_{lp}} + {\widehat G_{hp}} = \textbf{1}.
\end{eqnarray}
We also show the frequency response of the Gaussian complementary filter banks in Figures \ref{figure4}(e)--(f).

\begin{figure}
\begin{center}
   \subfigure[\scriptsize high frequency subspace]{\includegraphics[width=0.4\textwidth]{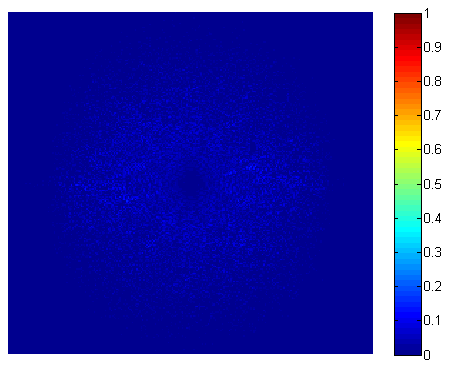}}
   \subfigure[\scriptsize low frequency subspace]{\includegraphics[width=0.4\textwidth]{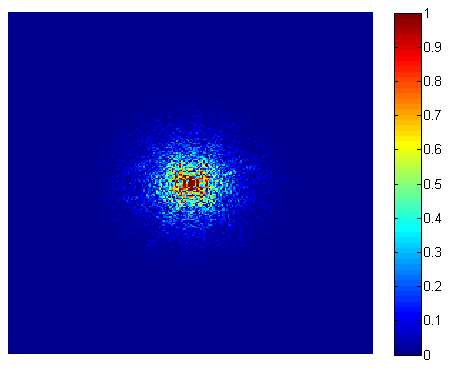}}
   \caption{The high and low frequency subspaces of a brain MRI obtained with the Gaussian filtering decomposition corresponding to Figures \ref{figure4}(e)--(f).  The sum of Figure \ref{figure5}(a) and \ref{figure5}(b) is equal to Figure \ref{figure1b}. However, we note that these are not simply disjoint partitions of the space.}
\label {figure5}
\end{center}
\end{figure}

To illustrate the proposed decomposition scheme, we show the brain MRI k-space magnitudes of the two subspaces using Gaussian subspace decomposition scheme in Figure \ref {figure5}. The sum of Figure \ref{figure5}(a) and \ref{figure5}(b) is equal to Figure \ref{figure1b}. However, as Figure \ref{figure5}(a) and \ref{figure5}(b) indicate, these subspaces do not simply correspond to disjoint partitions of k-space. Also in Figure \ref{figure5}(a), we observe the magnitudes within the high frequency subspace keep in the same range, which will benefit the k-space relative reconstruction accuracy.

\subsection{An equivalent objective}\label{sec.overall-objective}
We next briefly summarize the basic objective function that our DAC algorithm is optimizing. Recall that the typical objective function for CS-MRI has the form
\begin{equation}
 x = \arg\min_x \|y - F_u x\|_2^2 + \rho(x).
\end{equation}
In the algorithm described above, we modify this to
\begin{equation}
 x = \arg\min_x \sum_{i=1}^n \lambda_i\|H_i x-x_i\|^2_2 +  \min_{x_i} \|y_i - F_u x_i\|^2_2 + \rho_i(x_i).
\end{equation}
As can be seen, we minimize in two parts. First, we minimize $x_i$ over the two right-most terms in the ``divide'' portion of the algorithm. We then minimize over $x$ in the first term using the learned $x_i$. This way, the low and high frequency subspaces can contribute more equally to the reconstruction of $x$.

Finally, the setting of $\lambda_i$ is important when reconstructing the MRI $x$. We found that a uniform setting consistently works well. As another approach, viewing the first term as an augmented Lagrangian that attempts to enforce what is originally a strict equality $x_i = H_i x$, we also experiment with maximizing over $\lambda_i$ in an adversarial manner to try and enforce these equalities (subject to $\|\lambda\|_2 = 1$). We use this approach in our experiments. We summarize the entire procedure in Algorithm \ref{alg.1}.

\begin{algorithm}[t]
\caption{Divide-and-conquer CS-MRI}\label{alg.1}
 \begin{algorithmic}
  \STATE{Input sub-sampled k-space $y$ and Fourier basis $F_u$.}
  \STATE{Select subspace filters $H_1,\dots,H_n$.}
  \STATE{Select CS-MRI penalizations $\rho_1,\dots,\rho_n$.}
  \FOR{each $i$}
  \STATE Filter $y$ using the $i$th filter as in Equation (\ref{eq7}).
  \STATE Optimize $x_i = \arg\min_{x_i} \|y_i - F_u x_i\|_2^2 + \rho_i(x_i)$.
  \ENDFOR
  \WHILE{not converged}
  \STATE{Set $x = \arg\min_x \sum_{i=1}^n \lambda_i\|H_i x - x_i\|_2^2$ using Eq.\ (\ref{eq20}).}
  \STATE{Set $\lambda_i = \|H_i x - x_i\|_2^2$ and normalize to unit L2 length.}
  \ENDWHILE
  \RETURN{Vectorized reconstructed MRI $x$.}
 \end{algorithmic}
\end{algorithm}

\section{Experiments}

\begin{figure}[htb!]
\begin{center}
\subfigure[\tiny $40\%$ Cartesian mask]{\label {figure6a}%
\includegraphics[width=0.235\textwidth]{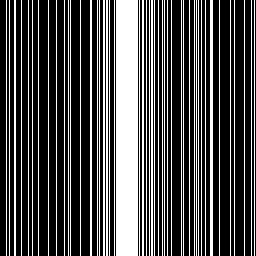}}
\subfigure[\tiny $30\%$ random mask]{\label {figure6b}%
\includegraphics[width=0.235\textwidth]{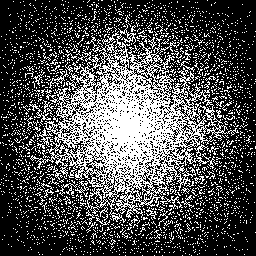}}
\subfigure[\tiny $30\%$ radial mask]{\label {figure6c}%
\includegraphics[width=0.235\textwidth]{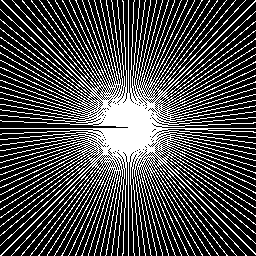}}\\
\subfigure[\tiny phantom data]{\label {figure6d}%
\includegraphics[width=0.235\textwidth]{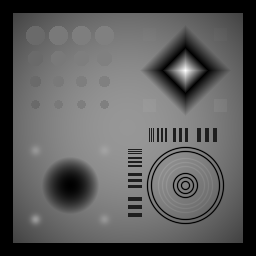}}
\subfigure[\tiny T2wBrain $27^{th}$ slice]{\label {figure6e}%
\includegraphics[width=0.235\textwidth]{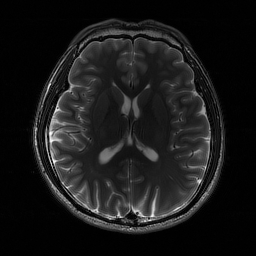}}
\subfigure[\tiny T2wBrain $7^{th}$ slice]{\label {figure6f}%
\includegraphics[width=0.235\textwidth]{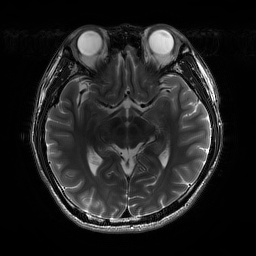}}
\subfigure[\tiny T1wBrain]{\label {figure6g}%
\includegraphics[width=0.235\textwidth]{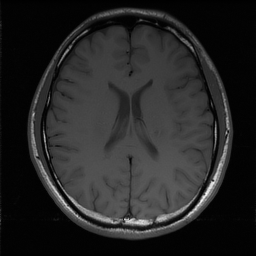}}\\
\caption{The MRI data and under-sampling masks used in our experiments.}
\label{figure6}
\end{center}
\end{figure}

\begin{figure}[htb!]
\begin{center}
\includegraphics[width=0.9\textwidth]{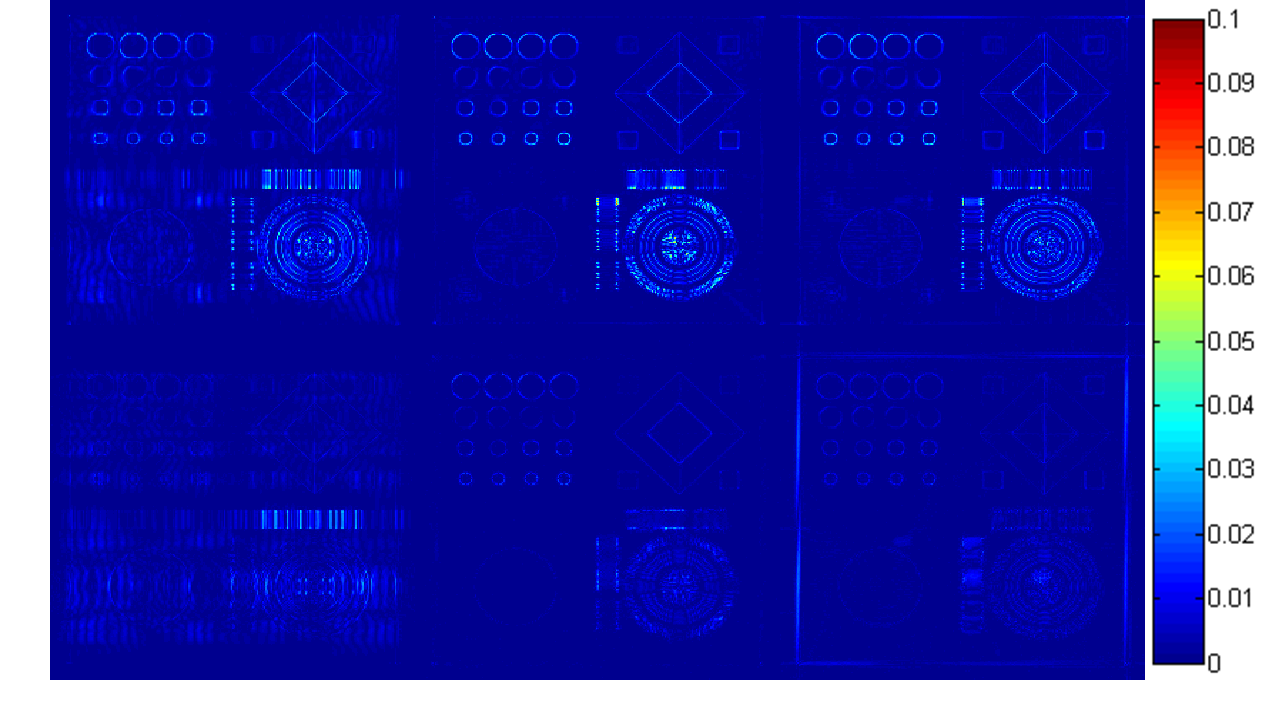}
\caption{The reconstruction errors results on phantom data using PBDWS and HoriVert PBDWS under three different under-sampling regimes. The first and second row represent the regular PBDWS and HoriVert PBDWS. From left to the right column we show the results with $40\%$ Cartesian mask, $30\%$ random mask and $30\%$ radial mask.}
\label{figure7}
\end{center}
\end{figure}

\subsection{Experimental setup}

For our experiments we adopt three sampling masks \cite{12}: 1D Cartesian sampling with random phase encodes, 2D random sampling, and pseudo radial sampling. These are shown in Figure  \ref{figure6}. The under-sampling ratio here means the ratio of k-space data being measured to the total number of full-sampled k-space data, it is negative correlated with reduction factor appears in other CS-MRI literatures. We conduct the simulations on the phantom shown in Figure \ref{figure6d} and the MRI-acquired complex-valued brain images shown in Figures \ref{figure6e}, \ref{figure6f} and \ref{figure6g}. These images are normalized to have maximum magnitudes of 1. As with other CS-MRI methods, the compressed data is acquired by simulating the under-sampling of the 2D DFT using the fully-sampled MRI.

We compare the Gaussian and HoriVert subspace filtering methods using the three state-of-the-art CS-MRI methods described in Section \ref{sec.reconstruction}. As performance measures we use the peak signal-to-noise ratio (PSNR), the structural similarity index (SSIM), and the high frequency error norm (HFEN) \cite{11}. The standard PSNR is a function of the MSE, but as we previously indicated, the PSNR measure is not the optimal choice in assessing the quality of an MR image. Therefore we also use SSIM and HFEN. SSIM measures the structural similarity of two images and is more consistent with the evaluation system of the human eye. HFEN has been proposed to evaluate the reconstruction quality of high frequency portions of MRI. In HFEN, the Laplacian of Gaussian (LoG) filter is used to extract the high frequency information within the MRI. HFEN is measured by the $l_2$ norm of the extracted features between the fully-sampled image and the reconstructed image.

All the experiments are coded in Matlab (R2014a). Computations are implemented with a Intel Core i5 CPU at 3.20GHz and 8G memory, employing a 64-bit Windows 7 operating system. For FCSA, WatMRI and PBDWS, we use the source code available from the authors' homepage, but we make parameter adjustments obtain the best performances.

\subsection{Illustrative experiment on phantom data}

The real-valued phantom image of the size $256\times256$ shown in Figure \ref{figure6d} is piece-wise constant and contains various image structures \cite{42}. The simluation phantom is created via Matlab implementation. Note that there exists rich low contrast, high frequency information in the phantom data. Thus this phantom data is more appropriate to evaluate the performance of the proposed DAC framework compared with conventional Shepp-Logan phantom. The Shepp-Logan phantom is extremely sparse under a gradient transform, so most CS algorithms can exactly reconstruct it from very few Fourier samples. To show the advantage of our divide-and-conquer method, we compare the reconstruction result of the original PBDWS algorithm with HoriVert PBDWS (DAC using HoriVert filters and PBDWS reconstruction) with a $40\%$ under-sampled Cartesian mask, $30\%$ under-sampled random mask and $30\%$ under-sampled radial mask.
In Figure \ref{figure7} we show the error residual images for each reconstruction. As is clear, the proposed DAC method is able to reconstruct the high frequency data more accurately by allowing the PBDWS algorithm to focus on these regions independently from the low frequency information. We again note that the same reconstruction algorithm is being used in both cases; the only difference is whether the sub-sampled k-space data is modeled directly or indirectly through different low and high pass filters.

\begin{figure}[htb!]
     \subfigure[Fully-sampled]{\label {figure8a}
     \includegraphics[width=0.235\textwidth]{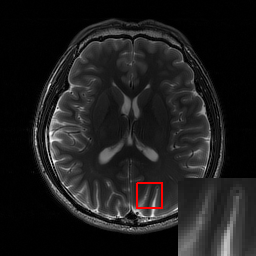}}
     \subfigure[FCSA]{\label {figure8b}
     \includegraphics[width=0.235\textwidth]{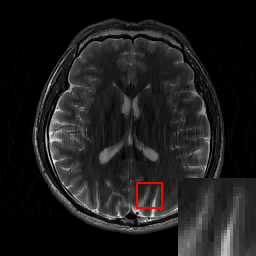}}
     \subfigure[Gau FCSA]{\label {figure8c}
     \includegraphics[width=0.235\textwidth]{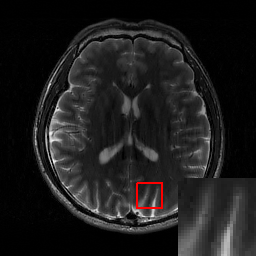}}
     \subfigure[HV FCSA]{\label {figure8d}
     \includegraphics[width=0.235\textwidth]{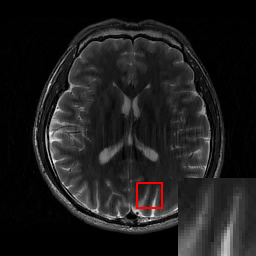}}\\
    \subfigure[Fully-sampled]{\label {figure8e}
    \includegraphics[width=0.235\textwidth]{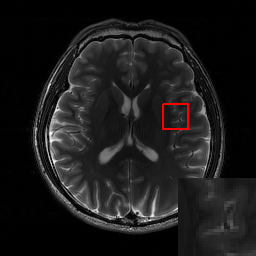}}
    \subfigure[FCSA]{\label {figure8f}
    \includegraphics[width=0.235\textwidth]{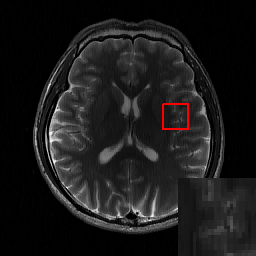}}
    \subfigure[Gau FCSA]{\label {figure8g}
    \includegraphics[width=0.235\textwidth]{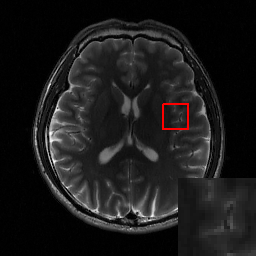}}
    \subfigure[HV FCSA]{\label {figure8h}
    \includegraphics[width=0.235\textwidth]{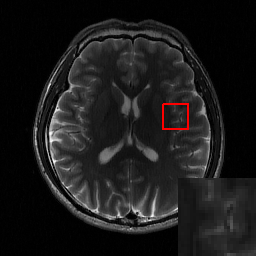}}\\
    \subfigure[Fully-sampled]{\label {figure8i}
    \includegraphics[width=0.235\textwidth]{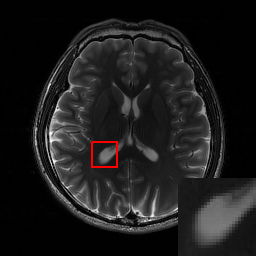}}
    \subfigure[FCSA]{\label {figure8j}\\
    \includegraphics[width=0.235\textwidth]{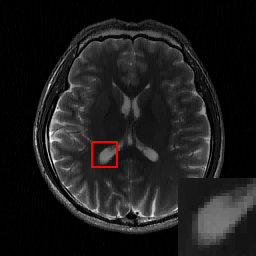}}
    \subfigure[Gau FCSA]{\label {figure8k}
    \includegraphics[width=0.235\textwidth]{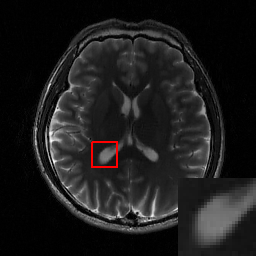}}
    \subfigure[HV FCSA]{\label {figure8l}
    \includegraphics[width=0.235\textwidth]{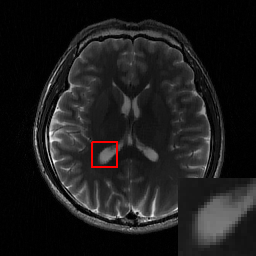}}\\
    \caption{The experiments conducted on T2wBrain slice27 data using FCSA method under Gaussian and HoriVert subspace schemes. The 1D $40\%$ under-sampling Cartesian mask,2D $30\%$ under-sampling random mask and $30\%$ under-sampling radial mask are applied in the first row, second row and third row experiments respectively. The reconstruction details are magnified in the red box.\newline}\label{figure8}
    \end{figure}

    \begin{figure}[htb!]
    \subfigure[Fully-sampled]{\label {figure9a}
    \includegraphics[width=0.235\textwidth]{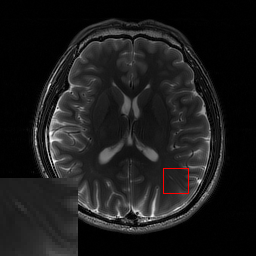}}
    \subfigure[PBDWS]{\label {figure9b}
    \includegraphics[width=0.235\textwidth]{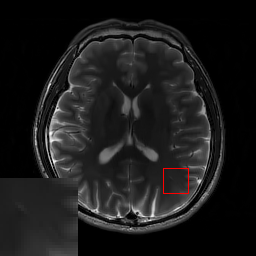}}
    \subfigure[Gau PBDWS]{\label {figure9c}
    \includegraphics[width=0.235\textwidth]{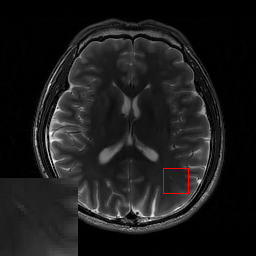}}
    \subfigure[HV PBDWS]{\label {figure9d}
    \includegraphics[width=0.235\textwidth]{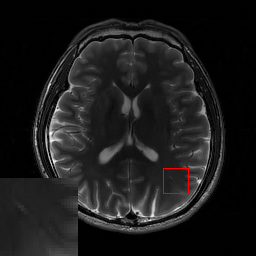}}\\
    \subfigure[Fully-sampled]{\label {figure9e}
    \includegraphics[width=0.235\textwidth]{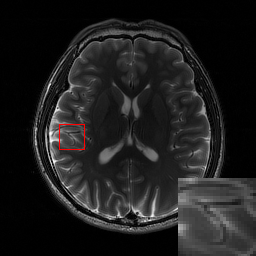}}
    \subfigure[PBDWS]{\label {figure9f}
    \includegraphics[width=0.235\textwidth]{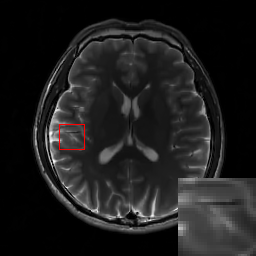}}
    \subfigure[Gau PBDWS]{\label {figure9g}
    \includegraphics[width=0.235\textwidth]{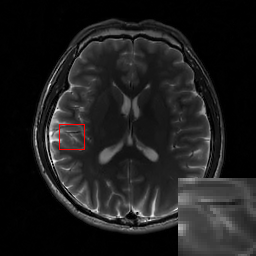}}
    \subfigure[HV PBDWS]{\label {figure9h}
    \includegraphics[width=0.235\textwidth]{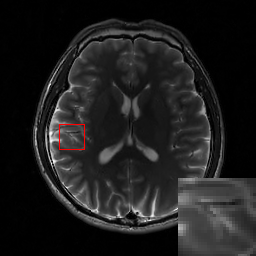}}\\
    \subfigure[Fully-sampled]{\label {figure9i}
    \includegraphics[width=0.235\textwidth]{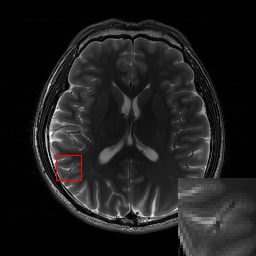}}
    \subfigure[PBDWS]{\label {figure9j}
    \includegraphics[width=0.235\textwidth]{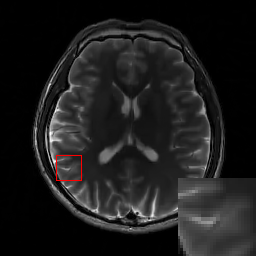}}
    \subfigure[Gau PBDWS]{\label {figure9k}
    \includegraphics[width=0.235\textwidth]{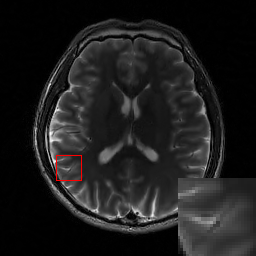}}
    \subfigure[HV PBDWS]{\label {figure9l}
    \includegraphics[width=0.235\textwidth]{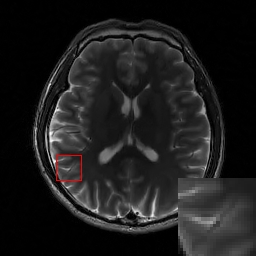}}\\
    \caption{The experiments conducted on T2wBrain slice27 data using PBDWS method under Gaussian and HoriVert subspace schemes. The 1D $40\%$ under-sampling Cartesian mask,2D $15\%$ under-sampling random mask and $15\%$ under-sampling radial mask are applied in the first row, second row and third row experiments respectively.} \label{figure9}
\end{figure}

\subsection{Experiments on T2wBrain data}

\begin{table*}[t!]\centering
 \tiny
 \caption{PSNR$|$SSIM$|$HFEN of the reconstruction for the T2wBrain 27slice MRI using various sampling masks and rates. Larger values are better for PSNR and SSIM, smaller values for HFEN.  In most cases divide-and-conquer improves the base algorithm.}
 \begin{tabular}{lcccc|ccccl}
  \toprule
  Mask & $\quad \% \quad$ & \,FCSA\, & \,Gauss FCSA\, & \,HoriVert FCSA \,& \,PBDWS \,& \,Gauss PBDWS \,& \,HoriVert PBDWS \,& \\
  \midrule
  Cartesian & 25 & {28.70}$|${0.850}$|${1.476} & 28.45$|$0.850$|$1.559 & {28.93}$|${0.869}$|${1.452} & 34.12$|$0.924$|$0.824 & {34.52}$|${0.944}$|${0.833} & {34.44}$|${0.950}$|${0.741}  \\
            & 30 & 31.71$|$0.893$|$1.042 & {32.54}$|${0.915}$|${0.964} & {32.48}$|${0.918}$|${0.968} & 36.79$|$0.940$|$0.558 & {37.81}$|${0.965}$|${0.544} & {37.19}$|${0.968}$|${0.483}
            \\
            & 35 & 32.16$|$0.903$|$0.944 & {33.08}$|${0.926}$|${0.888} & {33.06}$|${0.929}$|${0.884} & 37.58$|$0.945$|$0.511 & {38.67}$|${0.971}$|${0.487} & {38.18}$|${0.973}$|${0.435} \\
            & 40 & 33.33$|$0.914$|$0.758 & {35.03}$|${0.942}$|${0.671} & {34.80}$|${0.941}$|${0.668} & 39.05$|$0.951$|$0.420 & {40.43}$|${0.976}$|${0.395} & {39.71}$|${0.978}$|${0.355} \\

  Random    & 15 & 31.04$|$0.896$|$0.676 & {32.20}$|${0.922}$|${0.608} & {32.39}$|${0.927}$|${0.596} & 34.72$|$0.933$|$0.462 & {35.94}$|${0.960}$|${0.390} & {35.37}$|${0.961}$|${0.365} \\
            & 20 & 32.55$|$0.916$|$0.562 & {34.07}$|${0.944}$|${0.429} & {34.24}$|${0.945}$|${0.435} & 36.76$|$0.950$|$0.340 & {38.28}$|${0.971}$|${0.281} & {37.58}$|${0.972}$|${0.261} \\
            & 25 & 33.22$|$0.926$|$0.530 & {35.23}$|${0.952}$|${0.360} & {35.19}$|${0.950}$|${0.385} & 38.03$|$0.951$|$0.291 & {40.01}$|${0.978}$|${0.226} & {39.06}$|${0.979}$|${0.215} \\
            & 30 & 34.24$|$0.938$|$0.485 & {36.86}$|${0.963}$|${0.279} & {36.58}$|${0.959}$|${0.320} & 39.77$|$0.957$|$0.230 & {42.40}$|${0.984}$|${0.162} & {41.15}$|${0.984}$|${0.155} \\

  Radial    & 15 & {30.21}$|$0.879$|$1.057 & 30.19$|${0.878}$|${1.100} & {30.79}$|${0.901}$|${1.002} & 33.89$|$0.925$|$0.667 & {34.22}$|${0.947}$|${0.653} & {34.14}$|${0.953}$|${0.577} \\
            & 20 & 31.90$|$0.906$|$0.775 & {32.29}$|${0.908}$|${0.742} & {32.92}$|${0.925}$|${0.678} & 35.91$|$0.940$|$0.492 & {36.88}$|${0.965}$|${0.440} & {36.47}$|${0.968}$|${0.399} \\
            & 25 & 33.20$|$0.926$|$0.617 & {34.24}$|${0.930}$|${0.530} & {34.62}$|${0.940}$|${0.501} & 37.85$|$0.950$|$0.360 & {39.37}$|${0.976}$|${0.305} & {38.69}$|${0.977}$|${0.276} \\
            & 30 & 34.09$|$0.935$|$0.546 & {35.75}$|${0.943}$|${0.413} & {35.77}$|${0.947}$|${0.424} & 39.24$|$0.955$|$0.290 & {41.42}$|${0.981}$|${0.233} & {40.36}$|${0.982}$|${0.212} \\
  \bottomrule
 \end{tabular} \label {table1}
\end{table*}

\begin{figure}[htb!]
\begin{center}
    {\label {figure10}
    \includegraphics[width=1\textwidth]{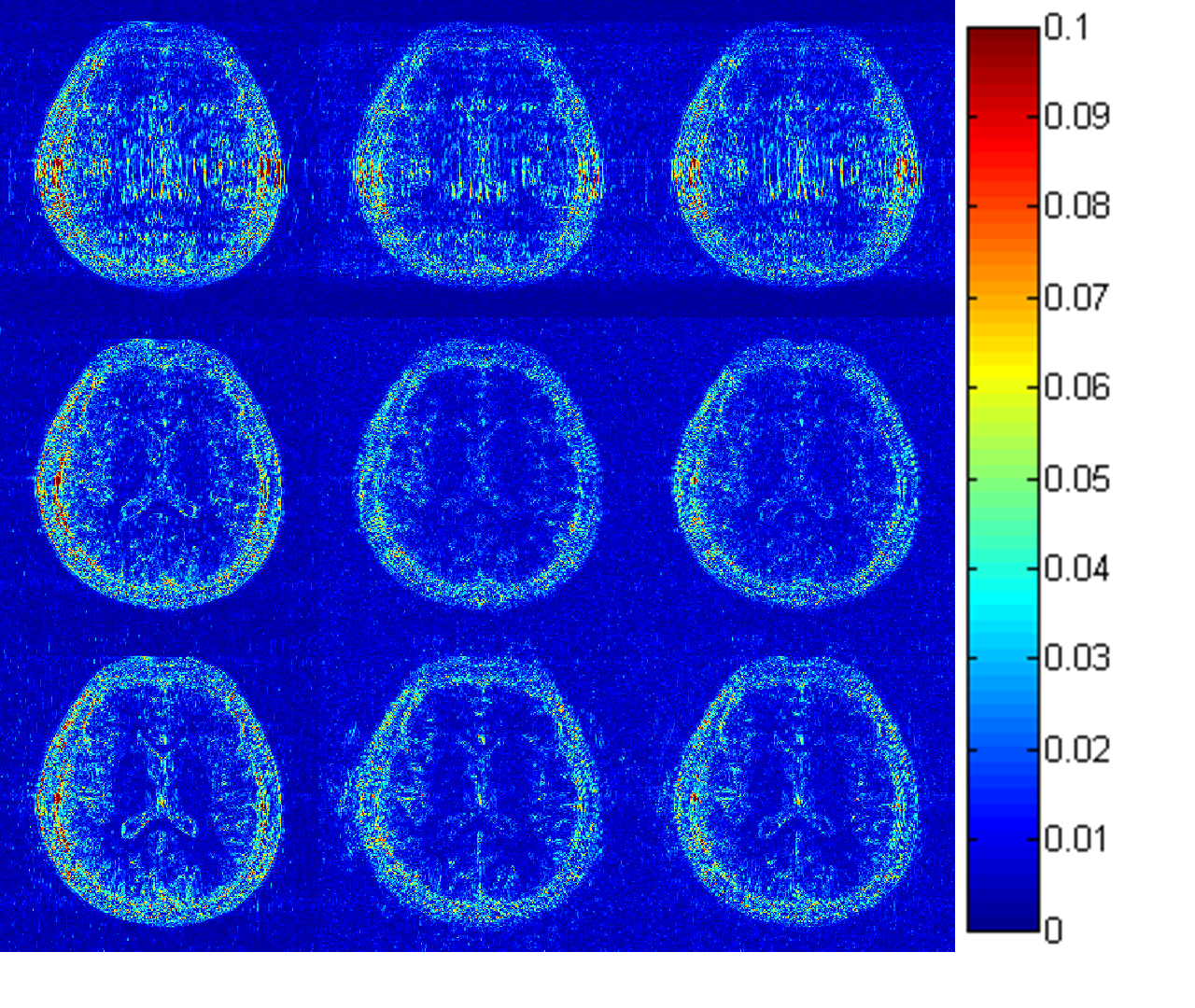}}\\
    \caption{The error residual images in Figure \ref{figure8}. The first row corresponds to Figure \ref{figure8b}, Figure \ref{figure8c}, Figure \ref{figure8d}. The second row corresponds to Figure \ref{figure8f}, Figure \ref{figure8g}, Figure \ref{figure8h}. The third row corresponds to Figure \ref{figure8j}, Figure \ref{figure8k}, Figure \ref{figure8l}.}
    \label {figure10}
\end{center}
\end{figure}

\begin{figure}[htb!]
\begin{center}
    {\label {figure11}
    \includegraphics[width=1\textwidth]{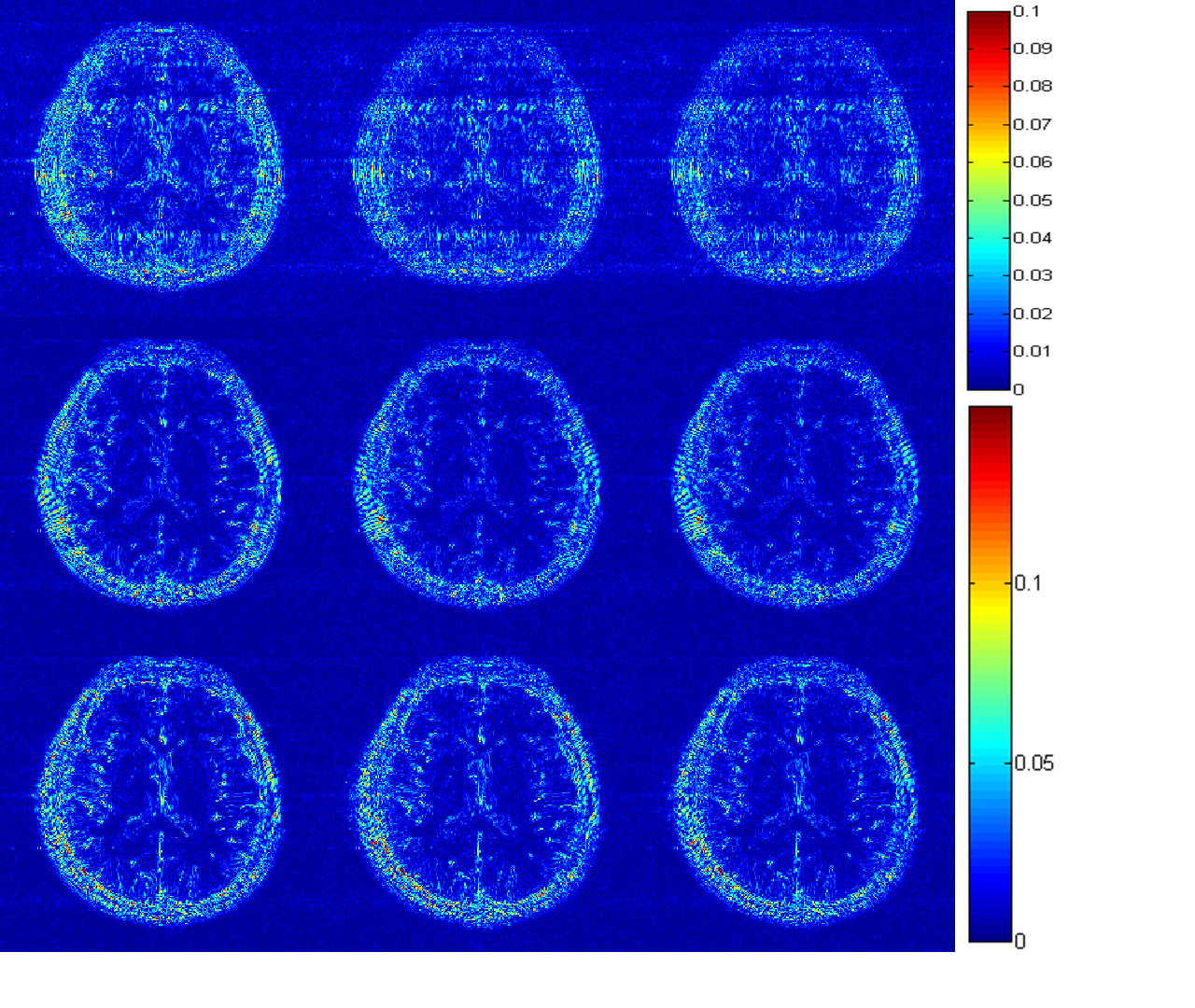}}\\
    \caption{The error residual images in Figure \ref{figure9}. The first row corresponds to Figure \ref{figure9b}, Figure \ref{figure9c}, Figure \ref{figure9d}. The second row corresponds to Figure \ref{figure9f}, Figure \ref{figure9g}, Figure \ref{figure9h}. The third row corresponds to Figure \ref{figure9j}, Figure \ref{figure9k}, Figure \ref{figure9l}. The first row corresponds to the error range from $0$ to $0.08$ while the second and third from $0$ to $0.15$.}
    \label {figure11}
\end{center}
\end{figure}

\begin{figure}[htb!]
\begin{center}
    {
    \includegraphics[width=1\textwidth]{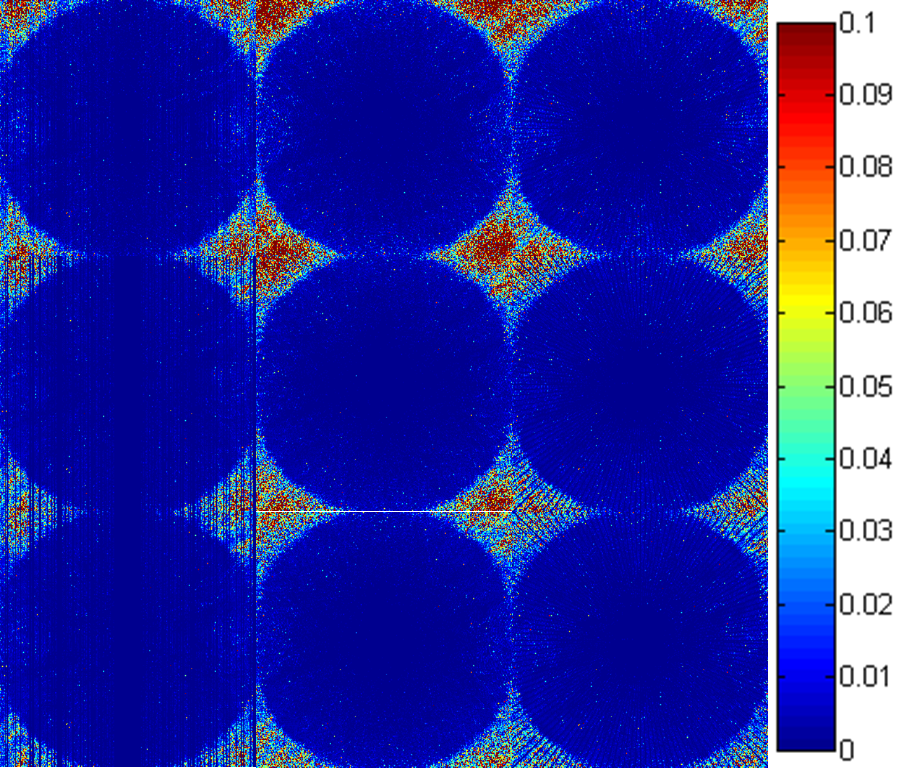}}\\
    \caption{The KRRE maps of the reconstructed images corresponds to the Figure \ref{figure10}. It is noted that the sampled positions in high frequency regions are reconstructed more accurately under the subspace framework.}
    \label {figure12}
\end{center}
\end{figure}

\begin{figure}[htb!]
\begin{center}
    {\label {figure13}
    \includegraphics[width=1\textwidth]{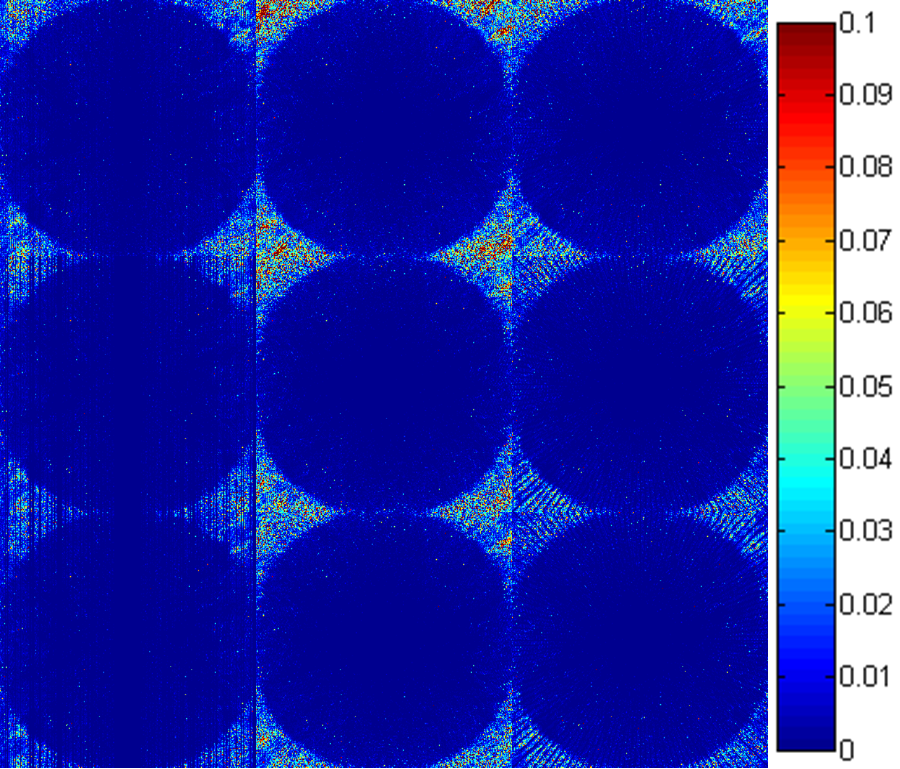}}\\
    \caption{The KRRE maps of the reconstructed images corresponds to the Figure \ref{figure11}.}
    \label {figure13}
\end{center}
\end{figure}

We also test our DAC framework on a clinically-obtained brain MRI also experimented with in \cite{6}, \cite{12}, \cite{15}, \cite{30} and \cite{31}. In particular, we use the $7^{th}$ and $27^{th}$ slices (named in the acquisition order) of a complex-valued T2-weighted brain MRI (size $256\times256$) volume data called T2wBrain (``slice7'' and ``slice27'' respectively, as shown in Figure \ref{figure6}), which is 2D acquired with 32 coils from a healthy volunteer by a 3-T Siemens Trio Tim MRI scanner using the T2-weighted turbo spin echo sequence (TR/TE=6100/99 $ms$, $220\times220$ $mm^2$ field of view, $3 mm$ slice thickness). We do SENSE reconstruction as the parallel imaging technique with reduction factor 1 to compose full k-space of gold standard images. The full k-space data will be used for emulate single-channel MRI.

We first test the T2wBrain slice27 data using Gaussian and HoriVert versions of FCSA and PBDWS.\footnote{As mentioned, we set the parameters to their best experimentally-obtained values according to SSIM index. For FCSA, these were $\mu = 2$, $\alpha = 0.003$ and $\beta = 0.001$ for the the high frequency subspace reconstruction and $\mu = 2$, $\alpha = 0.002$ and $\beta = 0.002$ for the low frequency subspace reconstruction. For PBDWS we set the data fidelity parameter $\mu=1e6$ for each subspace. We apply the same parameter setting for all the tested data. We detail the parameter setting in later discussion section.}
We show the reconstruction results for FCSA in Figure \ref{figure8}. The MR image details in the red box are magnified for better comparison. As is evident, for these sampling rates and patterns, the reconstructed images of FCSA have severe jagged visual effects because of poor high frequency reconstruction. These details are clearer using the proposed subspace method.

We also test with PBDWS and its DAC Gaussian and HoriVert extensions in Figure \ref{figure9}. Although PBDWS outperforms FCSA, high frequency degradation is similarly observable. With Gaussian and HoriVert PBDWS these fine structures are recovered better. We show the corresponding error residual images for both the FCSA and PBDWS based DAC framework in Figure \ref{figure10} and Figure \ref{figure11} respectively. The left column is the results directly using the CS-MRI method, the middle and right column correspond to these results with the Gaussian and HoriVert based DAC framework. The first row, second row and the third row corresponds to the Cartesian, random and radial under-sampling masks. The proposed subspace method shows smaller reconstruction errors in the structural details compared with the direct application of the same algorithms.

For further illustration, we plot the KRRE maps of the reconstructions in Figure \ref{figure12} and Figure \ref{figure13}. As can be seen, for our DAC method the high frequency regions of reconstructed k-space suffers less from errors than the direct method. This helps confirm our claim in Section \ref{sec.motivation} that isolating frequency content into subspaces for independent reconstruction allows for a more uniform reconstruction of k-space than the common squared error penalty.

The quantitative results for the $27^{th}$ slice of the T2wBrain data are given in Table \ref{table1}. As is clear, CS-MRI methods like FCSA and PBDWS can be significantly improved using the proposed divide-and-conquer method, which is consistent with the previous subjective evaluation. One interesting phenomenon is that, while the KRRE of Gaussian PBDWS is worse in high frequency regions than HoriVert PBDWS, the PSNR of Gaussian PBDWS is better than HoriVert PBDWS, while the SSIM and HFEN evaluation gives the opposite conclusion. This helps confirms the claim in Section \ref{sec.motivation} that the PSNR index does not provide a completely convincing measure of reconstruction quality in terms of detail recovery. SSIM and HFEN are designed to measure this, and these quantitative measures are more in agreement with the shown KRRE maps and subjective evaluation. From Figure \ref{figure11}, we observe here that the HoriVert subspace DAC framework slightly
outperforms the Gaussian counterparts in visual performance.

\begin{figure}[t!]
\begin{center}
    \subfigure[Fully-sampled]{\label {figure14a}
    \includegraphics[width=0.3\textwidth]{T2wBrain_slice7.png}}
    \subfigure[30\% radial mask]{\label {figure14b}
    \includegraphics[width=0.3\textwidth]{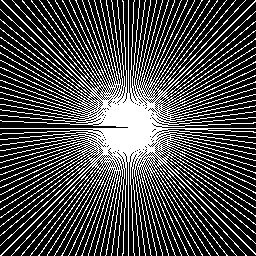}}
    \subfigure[20\% random mask]{\label {figure14c}
    \includegraphics[width=0.3\textwidth]{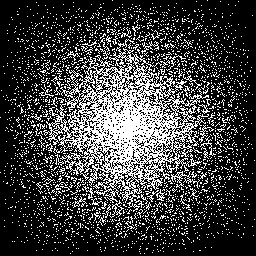}}\\
    \subfigure[WatMRI]{\label {figure14d}
    \includegraphics[width=0.3\textwidth]{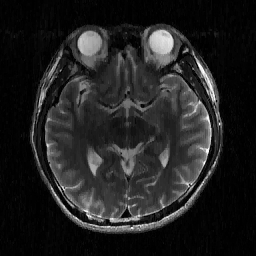}}
    \subfigure[Gau WatMRI]{\label {figure14e}
    \includegraphics[width=0.3\textwidth]{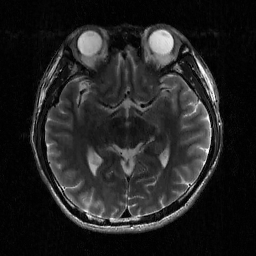}}
    \subfigure[HV WatMRI]{\label {figure14f}
    \includegraphics[width=0.3\textwidth]{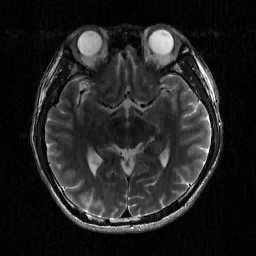}}\\
    \subfigure[PBDWS]{\label {figure14g}
    \includegraphics[width=0.3\textwidth]{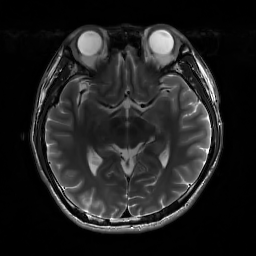}}
    \subfigure[Gau PBDWS]{\label {figure14h}
    \includegraphics[width=0.3\textwidth]{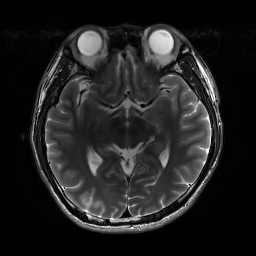}}
    \subfigure[HV PBDWS]{\label {figure14i}
    \includegraphics[width=0.3\textwidth]{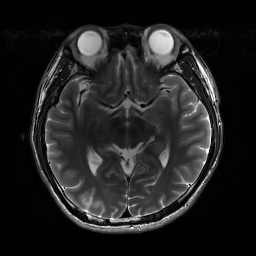}}\\
    \caption{The experiments conducted on T2wBrain slice7 data using WatMRI with $30\%$ radial under-sampling and PBDWS with $25\%$ random under-sampling under Gaussian subspace schemes. }
    \label {figure14}
\end{center}
\end{figure}

\begin{figure}[t!]
\begin{center}
    \subfigure{
    \includegraphics[width=1\textwidth]{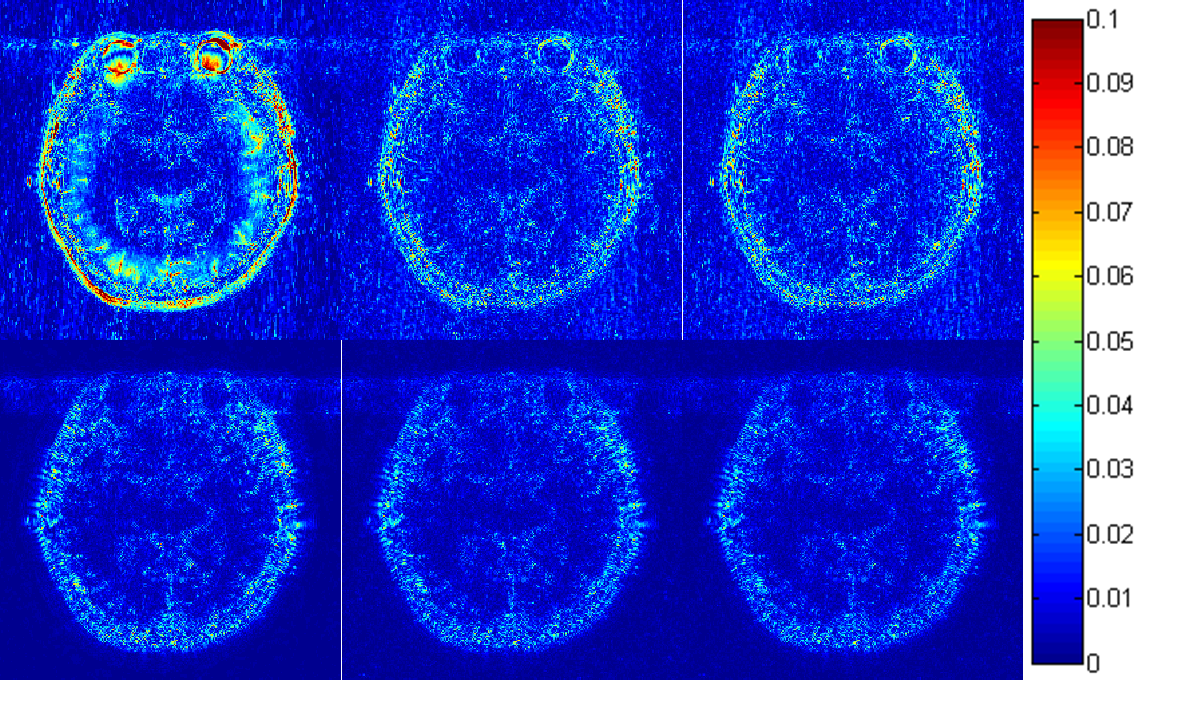}}
    \caption{The error residual images in Figure \ref{figure14}. The first row corresponds to Figure \ref{figure14d}, Figure \ref{figure14e}, Figure \ref{figure14f}. The second row corresponds to Figure \ref{figure14g}, Figure \ref{figure14h}, Figure \ref{figure14i}.}
    \label {figure15}
\end{center}
\end{figure}

Finally, we also conduct experiments on the T2wBrain slice7 using Gaussian WatMRI and Gaussian PBDWS, as shown in the Figure \ref {figure14}. We also give the error residual images in Figure \ref{figure15}. Again, Gaussian WatMRI and Gaussian PBDWS achieve better performance than their standard counterparts, WatMRI and PBDWS.

\subsection{Experiments on T1wBrain data}

In addition to the complex-valued in-vivo T2-weigthed brain MRI data, we also test on a complex-valued in-vivo T1-weighted brain MRI image called T1wBrain to validate the proposed framework on different MRI modalities \cite{14}. The T1wBrain image is an axial brain image from a 3T commercial scanner (GE Healthcare, Waukesha, WI) with an eight-channel head coil (In Vivo, Gainesville, FL) using a two-dimensional T1-weighted spin echo protocol (TE/TR = $11/700 ms$, $22 cm$ FOV, $10$ slices, $256 \times 256$ matrix). We test various CS-MRI algorithms on the T1wBrain data for comparison, including L1-ESPIRiT \cite{43}, pFISTA \cite{44}, PANO \cite{15}, PBDWS \cite{31}, GBRWT \cite{37} and the proposed DAC Gaussian PBDWS. Note that L1-ESPIRiT uses the parallel imaging technique, while we use a strategy similar to the T2wBrain data to emulate the single coil imaging for other algorithms. We have adjusted the parameters of these algorithms to their best performance in PSNR.

\begin{figure}[t!]
\begin{center}
    \subfigure[Fully-sampled]{\label {figure16a}
    \includegraphics[width=0.3\textwidth]{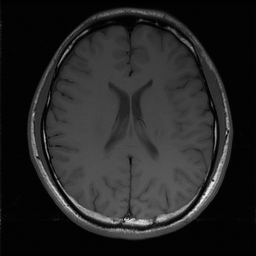}}
    \subfigure[Mask]{\label {figure16b}
    \includegraphics[width=0.3\textwidth]{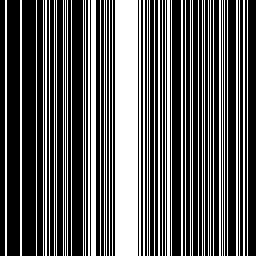}}\\
    \subfigure[L1-ESPIRiT]{\label {figure16c}
    \includegraphics[width=0.3\textwidth]{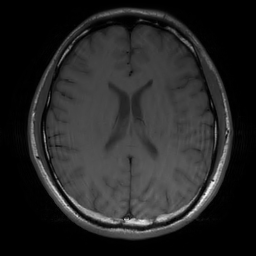}}
    \subfigure[pFISTA]{\label {figure16d}
    \includegraphics[width=0.3\textwidth]{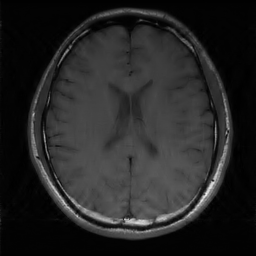}}
    \subfigure[PANO]{\label {figure16e}
    \includegraphics[width=0.3\textwidth]{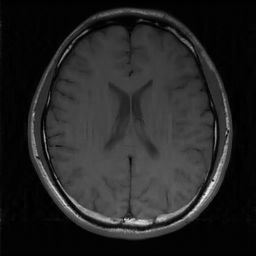}}\\
    \subfigure[PBDWS]{\label {figure16f}
    \includegraphics[width=0.3\textwidth]{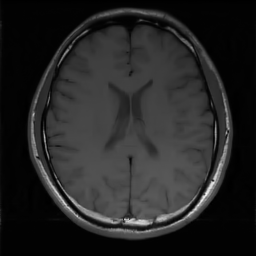}}
    \subfigure[GRBWT]{\label {figure16g}
    \includegraphics[width=0.3\textwidth]{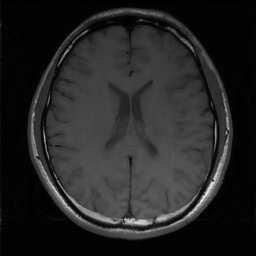}}
    \subfigure[Gau PBDWS]{\label {figure16h}
    \includegraphics[width=0.3\textwidth]{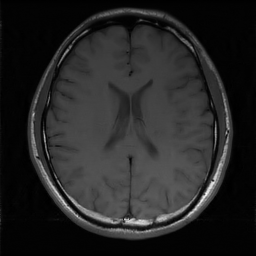}}
    \caption{Experiments conducted on T1wBrain data using various CS-MRI methods with $35\%$ Cartesian under-sampling. We note that L1-ESPIRiT is a parallel imaging CS-MRI algorithm.}
    \label {figure16}
\end{center}
\end{figure}

We give the reconstruction results and corresponding residual error images in Figures \ref{figure16} and \ref{figure17}. We see that structural information is preserved better under the DAC Gaussian PBDWS compared with other CS-MRI methods. To quantitatively assess these CS-MRI methods, we show the PSNR, SSIM and HFEN indexes in Figure \ref{figure18}. We observe that Gaussian PBDWS achieve the highest PSNR and SSIM value meanwhile the lowest HFEN value.

\begin{figure}[t!]
\begin{center}
    {\label {figure17}
    \includegraphics[width=1\textwidth]{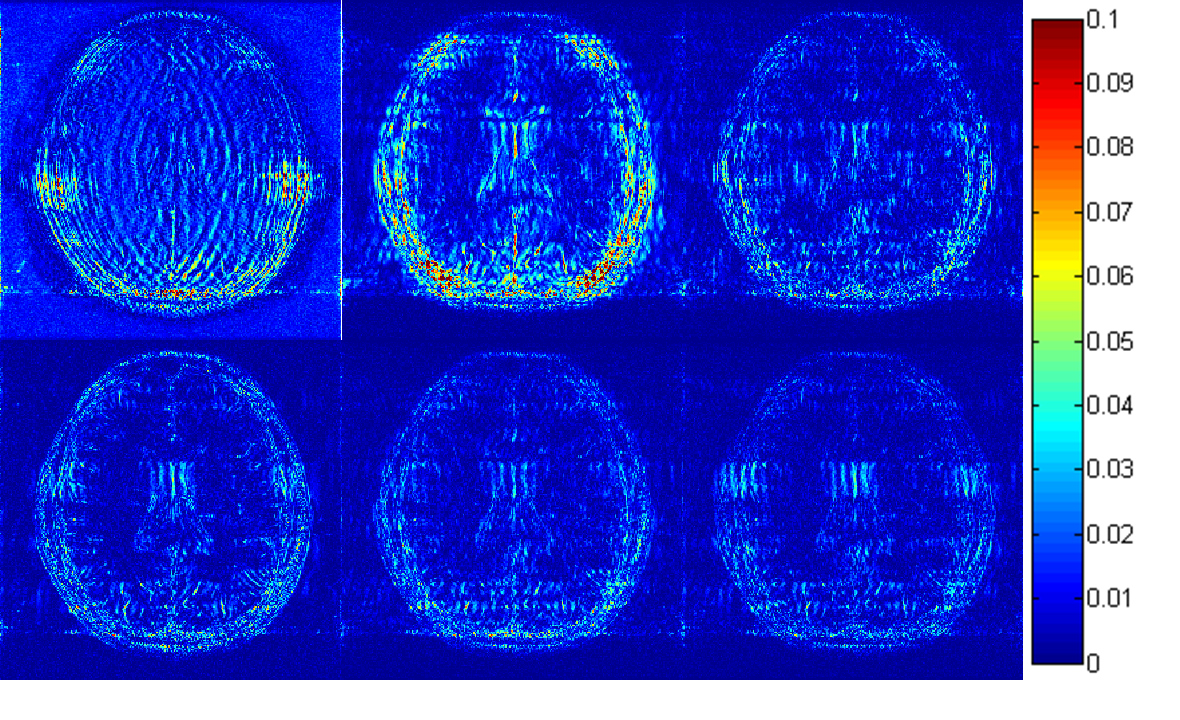}}
    \caption{The error residual images in Figure \ref{figure16}. We note the proposed DAC framework applied to the PBDWS algorithm using Gaussian subspace decomposition achieves the minimum reconstruction error. The first row corresponds to Figure \ref{figure16c}, Figure \ref{figure16d}, Figure \ref{figure14e} and The second row corresponds to Figure \ref{figure16f}, Figure \ref{figure16g}, Figure \ref{figure14h}.}
    \label {figure17}
\end{center}
\end{figure}

\begin{figure*}[t!]
\begin{center}
    {\label {figure18}
    \includegraphics[width=0.32\textwidth]{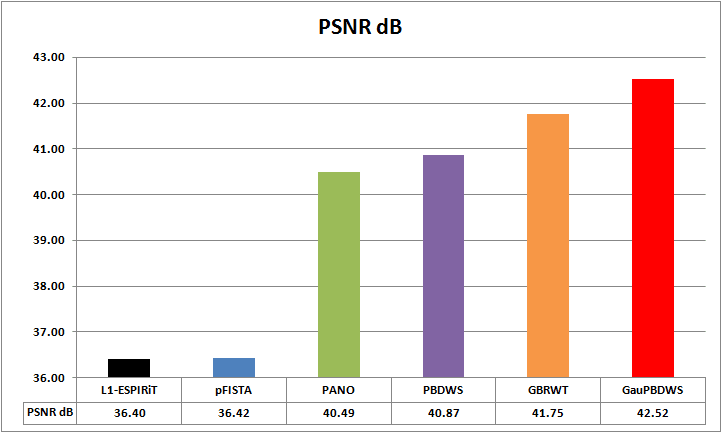}
    \includegraphics[width=0.32\textwidth]{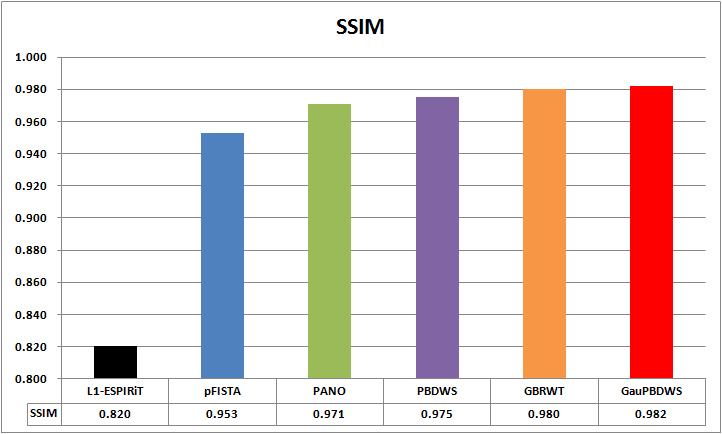}
    \includegraphics[width=0.32\textwidth]{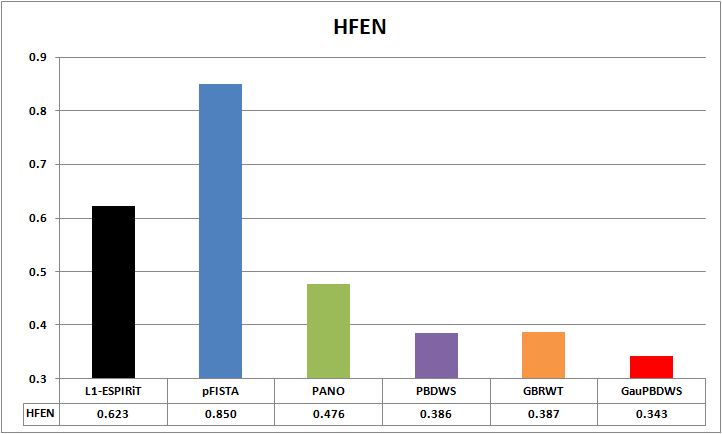}}
    \caption{The PSNR, SSIM and HFEN index for the experiments conducted on T1wBrain data with $35\%$ Cartesian under-sampling.}
    \label {figure18}
\end{center}
\end{figure*}

For the Gaussian and HoriVert subspace methods, the computational time required is roughly two and four times greater than the corresponding regular methods because there are two and four corresponding optimizations, respectively, rather than one. This constituted the most computationally intensive part of the proposed DAC framework, but we observe it is easily parallelizeable. For the subspace decomposition and subspace integration steps, the matrix Hadamard multiplication is computationally efficient.

\begin{figure}[t!]
\begin{center}
    \subfigure[PSNR]{\label {figure19a}
    \includegraphics[width=0.45\textwidth]{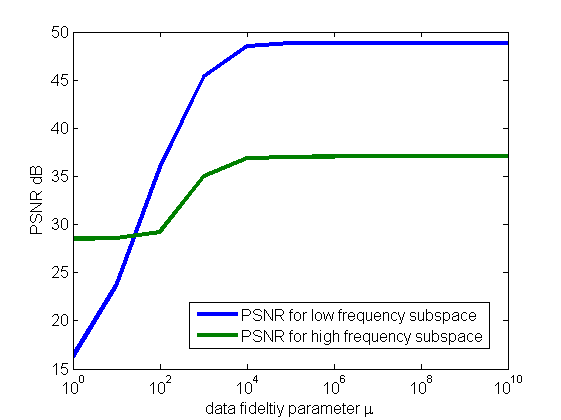}}
    \subfigure[SSIM]{\label {figure19a}
    \includegraphics[width=0.45\textwidth]{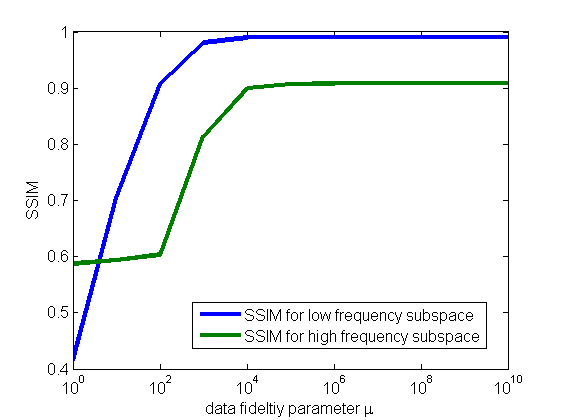}}
    \caption{The performance curve as a function of data fidelity regularization parameter $\mu$. The reconstruction remains stable when the regularization parameters reaches a certain value for both high and low frequency subspace.}
    \label {figure19}
\end{center}
\end{figure}

\section{Discussion}

\subsection{Discussion on parameter setting}

In the proposed divide-and-conquer framework, parameters requiring tuning are in both the subspace reconstruction and subspace integration stages. For subspace reconstruction, the number of parameters to be tuned only depends on the specific base algorithm adopted. If we divide k-space into $N$ subspaces, and the number of parameter for single subspace reconstruction is $P$, the total number of parameters for subspace reconstruction is $NP$. Therefore, if the chosen subspace reconstruction algorithm is robust to variations in parameter setting, the DAC extension of that algorithm will also be robust. For example, in the Gaussian PBDWS method the parameter to be adjusted is the data fidelity parameter $\mu_i$. For high and low frequency subspace reconstruction, we adjust the data fidelity regularization parameter low and high ranging from $1$ to $1e10$, and we plot the performance curve in PSNR and SSIM in Figure \ref{figure19}. The experiment is conducted using PBDWS on
T2wBrain $27^{th}$ slice with 2D $15\%$ mask.

We note the PSNR and SSIM index for both high and low frequency subspace reconstruction  reach the optimal around 1e4 and above, meaning for Gaussian PBDWS, when the regularization parameters exceeds 1e4, the method is not susceptible to them. We can choose an arbitrary regularization value greater than 1e4 for any data used by our DAC framework with PBDWS even if we have no access to the fully-sampled k-space data in real application scenarios. Hence we choose the regularization parameter in PBDWS used for subspace reconstruction to be 1e6, which is also recommended in the original paper in PBDWS.

For high and low frequency subspace reconstruction using FCSA/WatMRI, the regularization parameters for both the TV and wavelet terms can influence the subspace reconstruction, but we note that even when we use no parameter tuning in the subspace reconstruction phase, meaning the parameter setting is kept the same as the regular FCSA and WatMRI, the proposed method still outperforms these original methods by a considerable margin in Table \ref{table2}. This shows the improvement of the proposed DAC framework is not simply a result of parameter tuning.

\begin{table} [htbp] \center
 \caption{\label{table2} The experiments using same parameter setting.}
 \begin{tabular}{lcccl}
  \toprule
  & PBDW & Gaussian FCSA & HoriVert FCSA \\
  \midrule
  PSNR (dB) & 33.78 & 34.74 & 34.20 \\
  SSIM      & 0.914 & 0.922 & 0.915 \\
  \bottomrule
 \end{tabular}
\end{table}

%

As for the parameter setting in subspace integration, we find that model performance is already good by setting all the subspace integration parameters the same. These parameter can also be estimated via the proposed scheme in Algorithm 1 using the augmented Lagrangian method. In this way the strict equality of the subspace decomposition holds.

\subsection{Discussion on noisy environments}

During the acquisition of MRI measurements, the contamination brought by noise is inevitable.  Usually the SNR of the magnitude of a fully-sampled MRI image is the ratio between the mean of the magnitudes and the noise standard deviation estimated from the background. Taking the $27^{th}$ slice of T2wBrain data for example, the SNR of the fully-sampled is $37.97$. The SNR index for the high frequency subspace decreases because the noise is amplified. As discussed in Section II, the magnitudes of the high frequency subspace MRI are small yet important, because it contains structural information and fine details. With an efficient CS-MRI algorithm, we can denoise the high frequency subspace MR images while retaining image structures because CS-MRI algorithms can benefit from high
sparsity.

\begin{figure}[t!]
\begin{center}
    \subfigure[PSNR]{\label {figure20a}
    \includegraphics[width=0.45\textwidth]{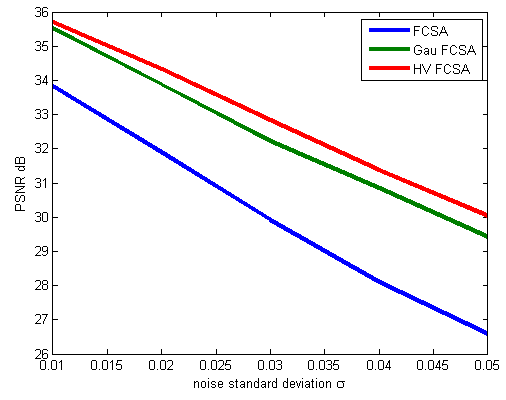}}
    \subfigure[SSIM]{\label {figure20b}
    \includegraphics[width=0.45\textwidth]{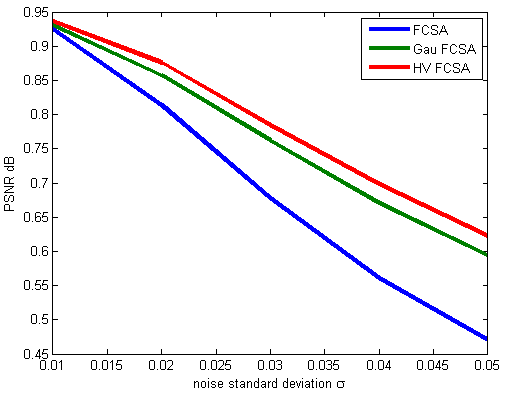}}\\
    \subfigure[PSNR]{\label {figure20c}
    \includegraphics[width=0.45\textwidth]{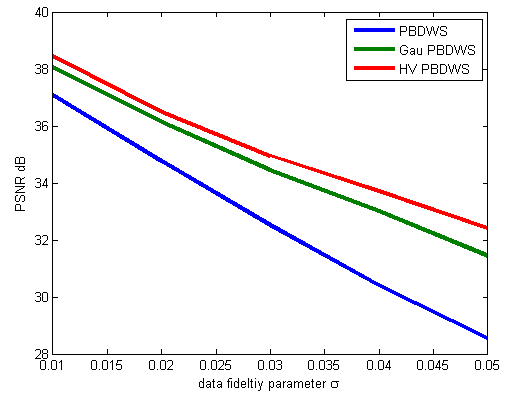}}
    \subfigure[SSIM]{\label {figure20d}
    \includegraphics[width=0.45\textwidth]{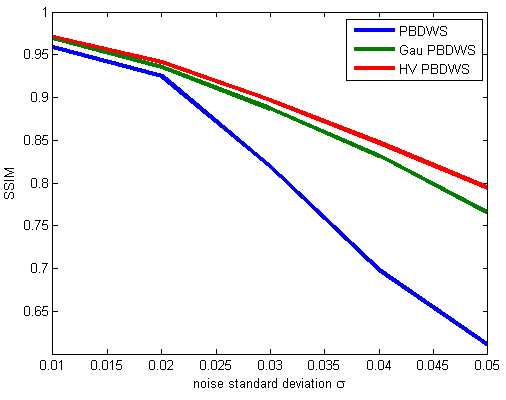}}\\
    \caption{Experiments conducted on T2wBrain slice27 data using FCSA, PBDWS and their DAC counterparts with $25\%$ random under-sampling. Zero-mean Gaussian noise is added with standard deviation ranging from $0.01$ to $0.05$.}
    \label {figure20}
\end{center}
\end{figure}

We have experimented with simulated noisy environments to evaluate the performance of the proposed DAC framework by adding the Gaussian noise into the under-sampled k-space. We conduct experiments on the $27^{th}$ slice of the T2wBrain image, where we use $25\%$ 2D random mask for under-sampling. We add Gaussian random noise to the k-space with various standard deviations from $0.01$ to $0.05$ to evaluate its robustness to noise using both FCSA and PBDWS methods. We plot the performance curve for PSNR and SSIM with respect to different noise level in Figure \ref{figure20}. The experiments show that the proposed DAC framework is robust to noise contamination in k-space. From Figures \ref{figure20b} and \ref{figure20d}, we observe that the proposed DAC framework also outperforms in SSIM, meaning the high frequency information still better reconstructed in the presence of noise, despite its larger relative magnitude in this region. We also observe the margin to which the DAC frameworks outperform the regular counterparts increases as the noise level goes up.

\section{Conclusion}

Based on the common observation that the energy and sparsity of k-space is non-uniformly distributed, we propose a divide-and-conquer framework for CS-MRI inversion. We first apply a series of linear filters to decompose the subsampled k-space measurements into separate frequency views called subspaces. For this we use two filtering schemes called HoriVert decomposition and Gaussian decomposition based on the linear-vertical and Gaussian filters. We then reconstruct the corresponding MRI in each subspace independently using any off-the-shelf CS-MRI inversion algorithm. We obtain the final reconstructed MRI by integrating all the reconstructed subspace images using Tikhonov regularization.

The experimental results on simulated phantom data and acquired complex-valued T2wBrain and T1wBrain MRI data show that the proposed subspace method can improve the performance of existing state-of-the-art CS-MRI methods considerably. We also observe that the proposed method has potential for recovering finer high-frequency details for diagnosis, which may improve the reliability and effectiveness of CS-MRI.

\section*{Acknowledgment}

This work was supported in part by the National Natural Science Foundation of China under Grants 61571382, 81671766, 61571005, 81671674, U1605252, 61671309 in part by the Guangdong Natural Science Foundation under Grant 2015A030313007, in part by the Fundamental Research Funds for the Central Universities under Grant 20720160075, 20720180059, in part by the National Natural Science Foundation of Fujian Province, China under Grant 2017J01126.

\section*{References}

\bibliography{mybibfile}

\end{document}